\documentclass[10pt,twocolumn,letterpaper]{article}

\usepackage[pagenumbers]{cvpr} %

\usepackage{graphicx}
\usepackage{amsmath}
\usepackage{amssymb}
\usepackage{color}
\usepackage{epsfig}
\usepackage{wrapfig,lipsum,booktabs}
\usepackage{enumitem}
\usepackage[normalem]{ulem}

\newcommand{\ra}[1]{\renewcommand{\arraystretch}{#1}}

\usepackage[pagebackref,breaklinks,colorlinks]{hyperref}

\usepackage[capitalize]{cleveref}
\crefname{section}{Sec.}{Secs.}
\Crefname{section}{Section}{Sections}
\Crefname{table}{Table}{Tables}
\crefname{table}{Tab.}{Tabs.}

\def\cvprPaperID{6712} %
\def\confName{CVPR}
\def\confYear{2022}

\begin{document}

\title{SpaceEdit: Learning a Unified Editing Space for Open-Domain Image Editing}

\author{Jing Shi$^{1}$\quad Ning Xu$^2$\quad Haitian Zheng$^1$\quad Alex Smith$^2$\quad Jiebo Luo$^1$\quad Chenliang Xu$^1$\\
$^1$University of Rochester \quad\quad $^2$Adobe Research
}

\twocolumn[{
\renewcommand\twocolumn[1][]{}%
\maketitle
\thispagestyle{empty}
\begin{center}
    \centering
    \includegraphics[width=\textwidth]{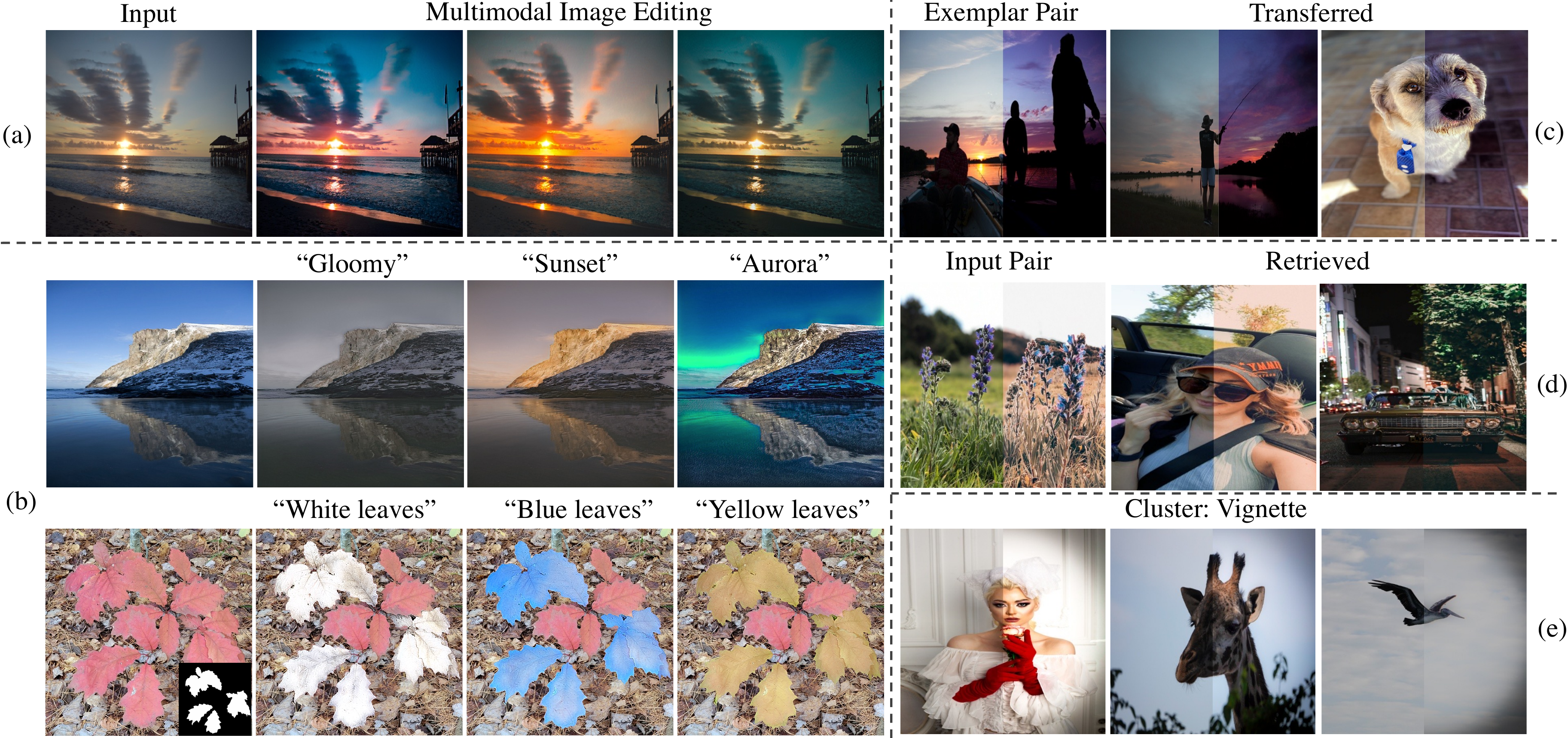}
    \captionof{figure}{We propose a new image editing paradigm with a unified model that can handle various open-domain image editing tasks: (a) multimodal image editing, (b) language-guided image editing, (c) examplar-based image editing, (d) editing style retrieval, (e) editing style clustering. Images in (c)-(e) are visualized as half-before half-after edited.}
    \label{fig:teasing}
\end{center}
}]

\begin{abstract}
Recently, large pretrained models (\eg, BERT, StyleGAN, CLIP) have shown great knowledge transfer and generalization capability on various downstream tasks within their domains. Inspired by these efforts, in this paper we propose a unified model for open-domain image editing focusing on color and tone adjustment of open-domain images while keeping their original content and structure. Our model learns a unified editing space that is more semantic, intuitive, and easy to manipulate than the operation space (\eg, contrast, brightness, color curve) used in many existing photo editing softwares. Our model belongs to the image-to-image translation framework which consists of an image encoder and decoder, and is trained on pairs of before- and after-images to produce multimodal outputs. We show that by inverting image pairs into latent codes of the learned editing space, our model can be leveraged for various downstream editing tasks such as language-guided image editing, personalized editing, editing-style clustering, retrieval, \etc. We extensively study the unique properties of the editing space in experiments and demonstrate superior performance on the aforementioned tasks.
\vspace{-6mm}
\end{abstract}

\section{Introduction}
\label{sec:intro}

Image editing has shown wide spectrum of applications in various scenarios including image retouching~\cite{he2020conditional,song2021starenhancer}, style transfer~\cite{yim2020filter,yoo2019photorealistic}, language-guided image editing~\cite{liu2020open, li2020manigan,shi2021learning,jiang2021language}, image harmonization~\cite{guo2021image}, colorization~\cite{zhang2016colorful}, etc.
However, the current research landscape independently studies these tasks on small and diverse datasets, underscoring the commonality of the image editing required for each task.
As such, the customized approach for one specific task is cumbersome to extend to other related tasks, and the bespoke model trained on a particular dataset has difficulty generalizing to out-of-domain samples.

The recent surge of general pretrained architectures for vision~\cite{dosovitskiy2020image,chen2021exploring} and vision+language~\cite{lu2019vilbert,radford2021learning} unifies different model structures for related tasks into common ones.
These unified models are first trained on some pretraining datasets and then either fine-tuned on specific datasets or directly applied in a zero-shot manner for different downstream tasks. Numerous studies have demonstrated that the generalization and knowledge transfer capability of the pretrained models are key to their success.   Here comes a natural question, \textit{is there any unified pretraining task or network architecture that we can leverage for the scope of image editing?}
One related work is styleGAN~\cite{karras2019style}, which is trained to generate realistic images for closed-domain categories such as faces, cats, and cars. Since then, a series of manipulation works~\cite{shen2020interpreting,voynov2020unsupervised,collins2020editing,wu2021stylespace,roich2021pivotal,xia2021tedigan} have been built upon styleGAN by inverting a given image to its latent space and then manipulating the latent code to generate a new image while keeping the generator intact. 

Despite being successful for closed-domain image editing, styleGAN has not been demonstrated to generate open-domain user photos which could contain various objects and complex scenes, therefore compromising its generalizability and application scenarios. 
In this paper, we are interested in one particular area of the open-domain image editing problem, \ie, apply some artistic styles to a given photo to achieve a different look while keeping its original content, structure, and texture. 
Although not covering all editing scenarios, the applications of our problem are already quite useful and broad for many photo editors and photographers. 
Indeed many commercial photo editing softwares such as Adobe Lightroom provide some predefined global and local editing operations (\eg, contrast, brightness, color curves) to solve this problem. However, their editing interfaces are not intuitive or convenient for many users, especially beginners, which we hope to mitigate with our newly proposed editing framework.

To achieve our goal, we propose a pretraining task that is useful for many editing downstream tasks. 
The pretraining task aims to transform a given before-edited image into an after-edited image with some artistic editing style controlled by some random noise vector. 
To learn the pretraining task, we first collect a new large-scale dataset with 60k pairs of before- and after-photos from the Lightroom Discover website\footnote{https://lightroom.adobe.com/learn/discover}. 
Then we propose a new image generator that appends the styleGAN as a decoder to an image encoder. 
The encoder features are inserted into the styleGAN decoder at different layers to preserve the details of the original image. 
The modulation modules and the mapping network of styleGAN are inherited to generate multimodal outputs.

After the generator is trained, we use a recent method SeFa~\cite{shen2021closed} to analyze the latent semantic directions as well as use some GAN inversion method~\cite{karras2020analyzing} to obtain the latent code in the $\mathcal{W}$ space given a pair of before and after images. 
We find that our $\mathcal{W}$ space has similar controllability and semantic disentanglement as the original styleGAN, but the meaning of the two $\mathcal{W}$s are entirely different. 
The $\mathcal{W}$ space of styleGAN contains the complete content information of the generated images while our $\mathcal{W}$ space only captures various editing styles, which are independent of image content. 
We verify that our inverted latent code are useful for both generation and recognition (\eg clustering, retrieval) tasks.

Given the unique properties of our  editing space $\mathcal{W}$, we apply our pretrained generator to several open-domain image editing tasks. 
First, we explore the task of language-guided image editing (LGIE)~\cite{shi2021learning,jiang2021language}, which aims to edit an image to match a given editing request. 
Existing methods must train their full models with sophisticated pixel-level losses on the limited dataset, thus facing the overfitting issue given the enormous language and image space. 
In contrast, we propose a simple encoder which maps the input image and text features into the 512-dimensional editing space and then resorts to our pretrained generator to generate the output image. 
Experimental results verify the advantage of our pretrained model serving for this downstream task.

Second, inspired by recent styleCLIP~\cite{patashnik2021styleclip}, we further equip our generator with CLIP~\cite{radford2021learning} for zero-shot free-form LGIE. 
Our method is able to not only generate semantic editing styles such as ``sunset," ``gloomy," but also change the color of an object to different colors as shown in Fig.~\ref{fig:teasing}.

Last but not least, since each latent code of a before- and after-pair in $\mathcal{W}$ space corresponds to some editing style, we can transfer the editing style of one image pair to the other images to achieve personalized editing. 
Besides, we can retrieve similar editing styles for personal style recommendation on a large database of user editing examples.

In summary, our contributions are three-fold.
First, we propose a new pretraining task and a network architecture that is beneficial for various pertinent tasks for open-domain image editing.
Second, we demonstrate that the $\mathcal{W}$ space of the pretrained model corresponds to various editing styles. 
Such embeddings are useful for both generative and recognition tasks.
Finally, we demonstrate the efficacy of our pretrained model on various downstream tasks, including the state-of-the-art performance on multimodal image editing and language-guided image editing benchmarks.

\section{Related Work}
\label{sec:related_work}

\noindent\textbf{Leveraging GAN latent space  for image editing.}
Many works have been proposed to discover the semantics in GAN's latent space for image editing in the supervised way~\cite{goetschalckx2019ganalyze,shen2020interpreting},  self-supervised way~\cite{jahanian2019steerability,plumerault2020controlling}, and unsupervised way~\cite{voynov2020unsupervised,collins2020editing,wang2021geometry,shen2021closed,wu2021stylespace}.
However, all the above works focus on unconditional GANs while our method relies on   conditional GAN.
Although traversing the latent space of unconditional GANs can achieve image editing in closed-domain images such as faces, its incapability of generating real-world images (\eg, multiple objects and complex scenes) limits their generalization and application. In addition, since their hidden spaces need to retain all the information of the generated outputs, the inversion ~\cite{zhu2016generative} of an open-domain image is usually compromised for photo fidelity~\cite{abdal2020image2stylegan++,roich2021pivotal}. In contrast, the editing space of our proposed model does not have such limitations.
Moreover, since each inverted latent code in the editing space corresponds to some editing style, we can directly cluster them to find representative semantics, which is not investigated by previous methods.

\noindent\textbf{Multimodal image editing.}
Our pretraining is a multimodal image editing task which requires diverse outputs controlled by some random vectors given an input image. 
A branch of works achieves the multimodal diversity by using an inverse mapping from the generated image to the input noise~\cite{zhu2017multimodal}, disentangling of image content and style~\cite{huang2018multimodal,lee2018diverse}, or explicitly enforcing the image diversity with distance-based loss term~\cite{mao2019mode,liu2021divco}.
However, the enforcement of diversity deteriorates the image quality. Inspired by 
the recent modulation approach~\cite{zhao2021large} for multimodal image inpainting, we propose a similar network architecture specifically for open-domain image editing. The difference is that our modulation layer does not use the features of the input image, which leads to better fidelity and diversity.

\noindent\textbf{Language-guided image editing.}
Language is a flexible and user-friendly way to control image editing.
\cite{chen2018language,el2019tell,shi2020benchmark,shi2021learning,jiang2021language} collect paired data (\ie input image, language request, target image) for supervised training. However, the language annotation is expensive to obtain, and the limited data size would constrain the generalizability of the approaches.
Other works \cite{dong2017semantic,nam2018text,mao2019bilinear,li2020manigan,yu2019multi} are trained with only image caption pair but are restricted to domain-specific images such as birds and flowers. 
Recently, some attempts are made to achieve zero-shot open-vocabulary image editing~\cite{xia2021tedigan,bau2021paint,patashnik2021styleclip} by modifying the latent space of a pretrained StyleGAN~\cite{karras2019style} via a state-of-the-art image-text matching model CLIP~\cite{radford2021learning}.
Hence, the data domain that the styleGAN is pretrained on will limit the editing domain.
Although \cite{liu2020open} trains a generator by reconstruction and thus can work for any open image domain, the generation quality is not guaranteed.
In contrast, the editing quality of our method is guaranteed by the unique properties of our learned editing space. We propose different approaches for both supervised and zero-shot language-guided image editing. Each of them achieves better editing results than other state-of-the-art methods.

\begin{figure}[t]
	\centering
	\includegraphics[width=1\linewidth]{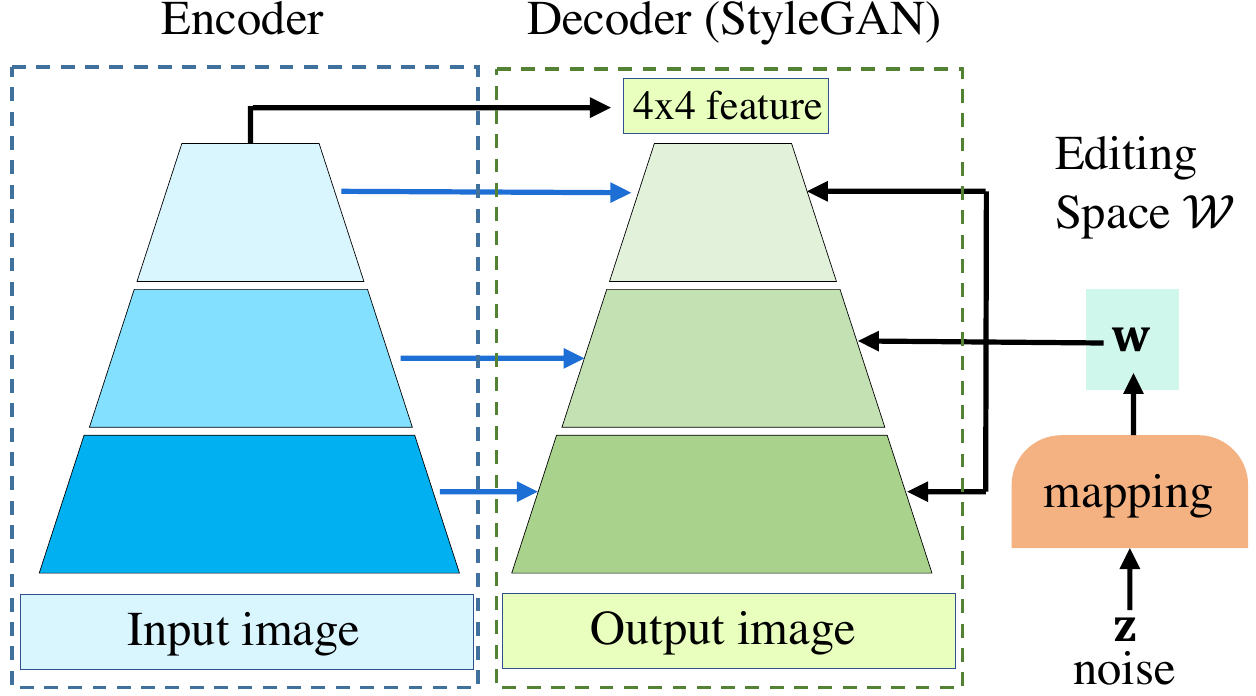}
    \caption{The structure of our generator for the pretraining task. The blue arrows represents skip connections.}
    \label{fig:main_structure}
\vspace{-4mm}
\end{figure}

\section{Multimodal Image Editing as Pretraining}
\label{sec:method}
For the pretraining task, our goal is to learn an image-conditional generator with a latent space that can control various editing styles.
The latent space should be semantic, disentangled as well as complete to be useful for various downstream editing tasks. 
We select multimodal image editing as our pretraining task as it encourages to produce diversified outputs with different editing styles. 

We propose an image-to-image translation framework that consists of an image encoder and an image decoder with some random noise $\mathbf z\in\mathcal{Z}$ as additional inputs to control different editing styles.
Since StyleGAN2~\cite{abdal2020image2stylegan++} has shown great disentanglement of its latent space for generative tasks, we adopt its architecture as our decoder where the noise input $\mathbf z$ is firstly mapped to an intermediate latent code $\mathbf w\in\mathcal{W}$, and then is further used to modulate the convolutional kernel at different layers, as depicted in Fig.~\ref{fig:main_structure}. The role of the image encoder is to encode the input image into features of different levels, and the lowest  4x4 feature map is used to replace the original constant input of StyleGAN2. 
Apart from the straightforward docking of the encoder and decoder, we further stitch them via skip connection at different resolutions of the feature maps from the encoder to decoder, in view of preserving fine-grained details.
Please refer to Appx.~\ref{appx:network_structure} for detailed structure.

More formally, let the source (before) image be $I_{in}$, the target (after) image $I_{tgt}$, the generator $G$, the discriminator $D$, the output image $I_{out}=G(I_{in}, \mathbf{w})$ where $\mathbf{w}=\mathrm{Mapping}(\mathbf{z})$. Our generator is trained with the regular conditional discriminator loss $\mathcal{L}_{adv}$ as
\begin{align}
    \mathcal{L}_{adv} =& -\mathbb{E}_{I_{in},I_{tgt}}[\log (D(I_{in}, I_{tgt}))] \nonumber\\
        =& -\mathbb{E}_{I_{in},I_{out}}[\log (1-D(I_{in}, I_{out}))].
\end{align}

\begin{figure}[t]
	\centering
	\includegraphics[width=1\linewidth]{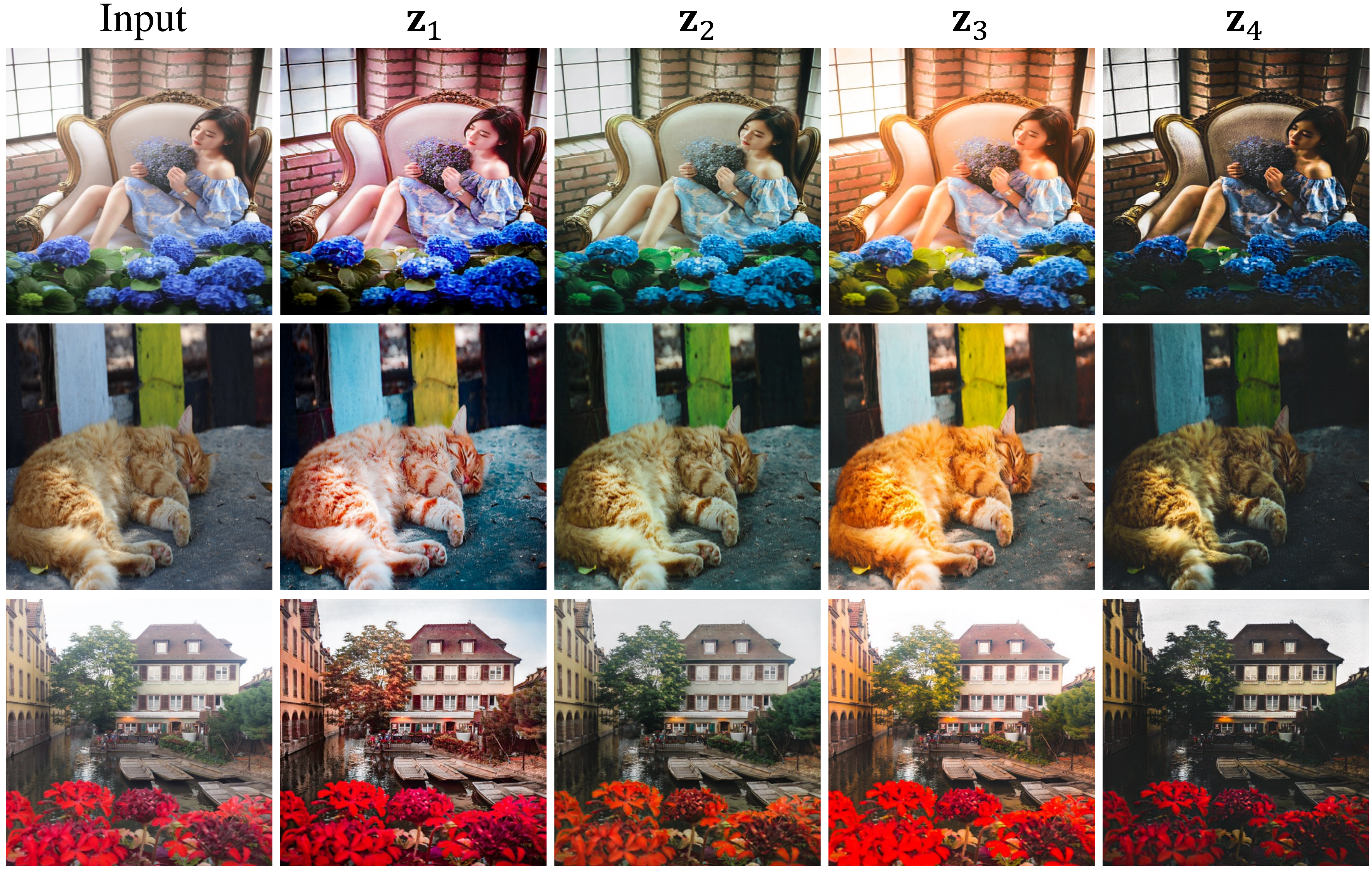}
    \caption{The multimodal image editing results controlled by different $\mathbf z$, each of which portrays one unique editing style. }
    \label{fig:viz_style}
\vspace{-3mm}
\end{figure}

Note that we circumvent direct pixel supervision such as L1 loss~\cite{isola2017image} for the purpose of encouraging the generation diversity, as suggested in ~\cite{zhao2021large}. Some qualitative output results from our trained generator is visualized in Fig.~\ref{fig:viz_style}. Our generator is  able to not only generate diverse outputs given different noise inputs on a single image, but also produce consistent editing styles given the same noise input on different images, indicating the independence between the learned editing space and image content.

\begin{figure}[t]
	\centering
	\includegraphics[width=1\linewidth]{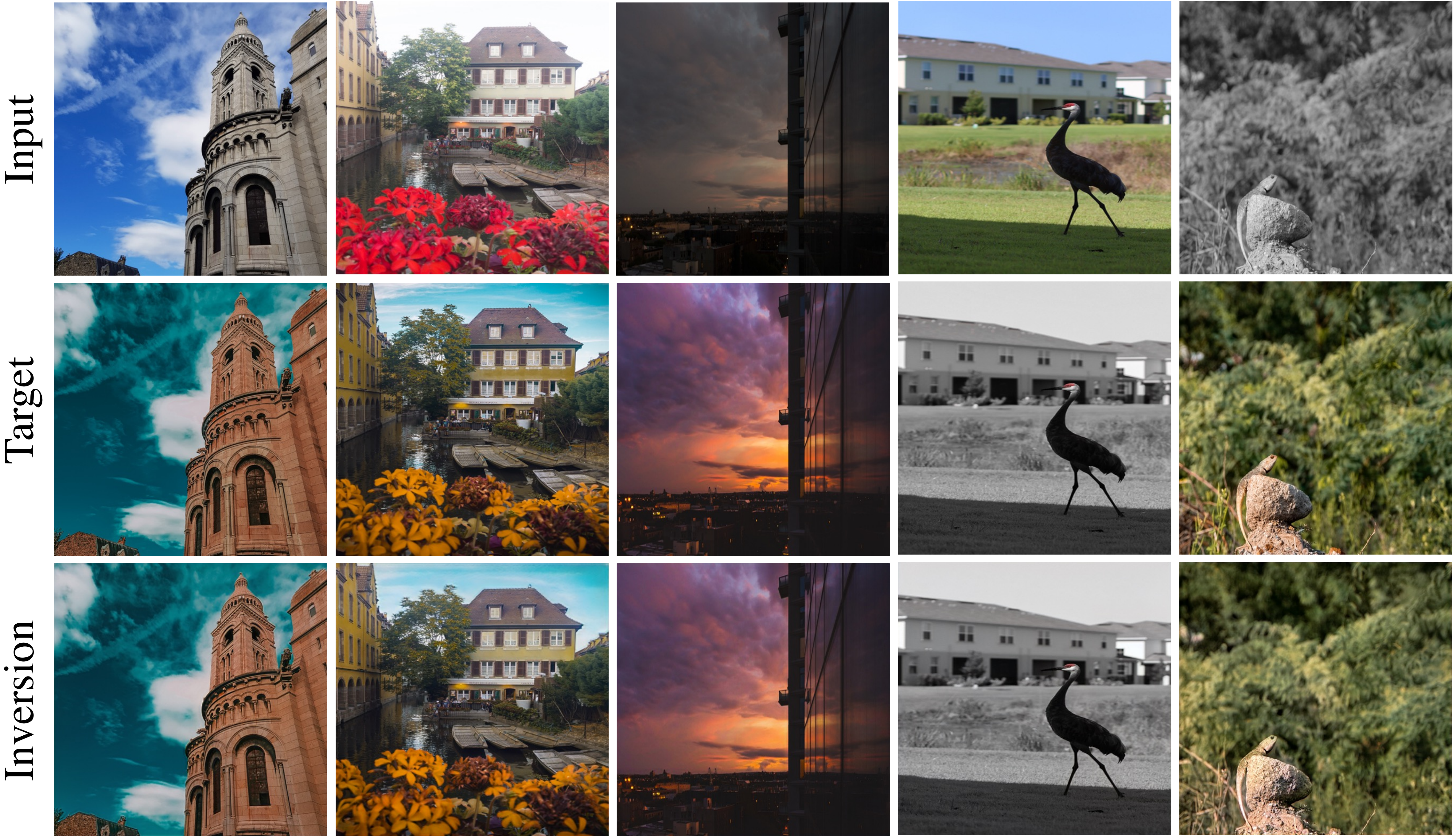}
    \caption{The visualization of conditional GAN inversion.}
    \label{fig:viz_inverse}
\vspace{-6mm}
\end{figure}

\section{Editing Space Analysis}

\subsection{Editing Space Inversion}
Similar to StyleGAN, the $\mathcal{W}$ space of our generator is more disentangled than the input $\mathcal{Z}$ space. Therefore we rely on the $\mathcal{W}$ space as the editing space for our editing tasks. 
The first question is whether the style embedding for any source and target image pair can be inverted into editing space, which measures the completeness and upper-bound editing ability of the $\mathcal{W}$ space.
To answer this question, we propose a \emph{conditional GAN inversion} problem: finding a $\mathbf w$ that can transfer the source image $I_{in}$ to the target $I_{tgt}$. We adapt an existing unconditioned GAN inversion method~\cite{zhu2016generative} to solve this problem, as formulated in Eq.~(\ref{eqn:condGAN_inv})
\begin{equation}\label{eqn:condGAN_inv}
    \mathbf w, \mathbf n = \arg\min_{\mathbf w,\mathbf n} \mathcal{L}_\mathrm{LPIPS}(I_{tgt}, G(I_{in},\mathbf w, \mathbf n)) + \lambda_n\mathcal{L}_n(\mathbf n),
\end{equation}
where $\mathbf{w}$ and $\textbf{n}$ are the inverted latent code and stochastic noise inputs to different layers of the decoder, respectively.  $\mathcal{L}_\mathrm{LPIPS}$ is the LPIPS perceptual loss~\cite{zhang2018unreasonable} and $\mathcal{L}_n$ denotes the noise regularization term~\cite{karras2020analyzing} with $\lambda_n$ as a balance weight. We show some randomly picked inversion results in 
Fig.~\ref{fig:viz_inverse}. It is clear that
our editing space $\mathcal
W$ can represent diverse editing styles such as drastic color manipulation, colorization, and local editing, which are useful for various downstream tasks. Besides qualitative results, we also show the quantitative result of reconstruction errors on both training and testing datasets in Tab.~\ref{tab:condGAN_inv}. 
\begin{wraptable}{r}{0.23\textwidth}
\vspace{-6mm}
\ra{1}
\center
\scalebox{1}{
\begin{tabular}{@{}crr@{}}
\toprule
Inversion & Train & Test \\
\midrule
Init & 24.88 & 24.93 \\
$\textbf w$ & 4.43  & 4.43 \\
$\textbf{w}_0$ & 1.86 & 1.86 \\
\bottomrule
\end{tabular}}
\caption{
Init, $\textbf w$, $\textbf w_0$ measure the \emph{mean pixel absolute error (maximum 255)} between source and target image, inverted and target image, source and reconstructed source image, respectively.}
\vspace{-8mm}
\label{tab:condGAN_inv}
\end{wraptable}
With inverted w, the outputs from our generator can almost reconstruct the target images perfectly with negligible $\sim$4 pixel errors, indicating the completeness of our learned editing space.

\begin{figure}[t]
	\centering
	\includegraphics[width=1\linewidth]{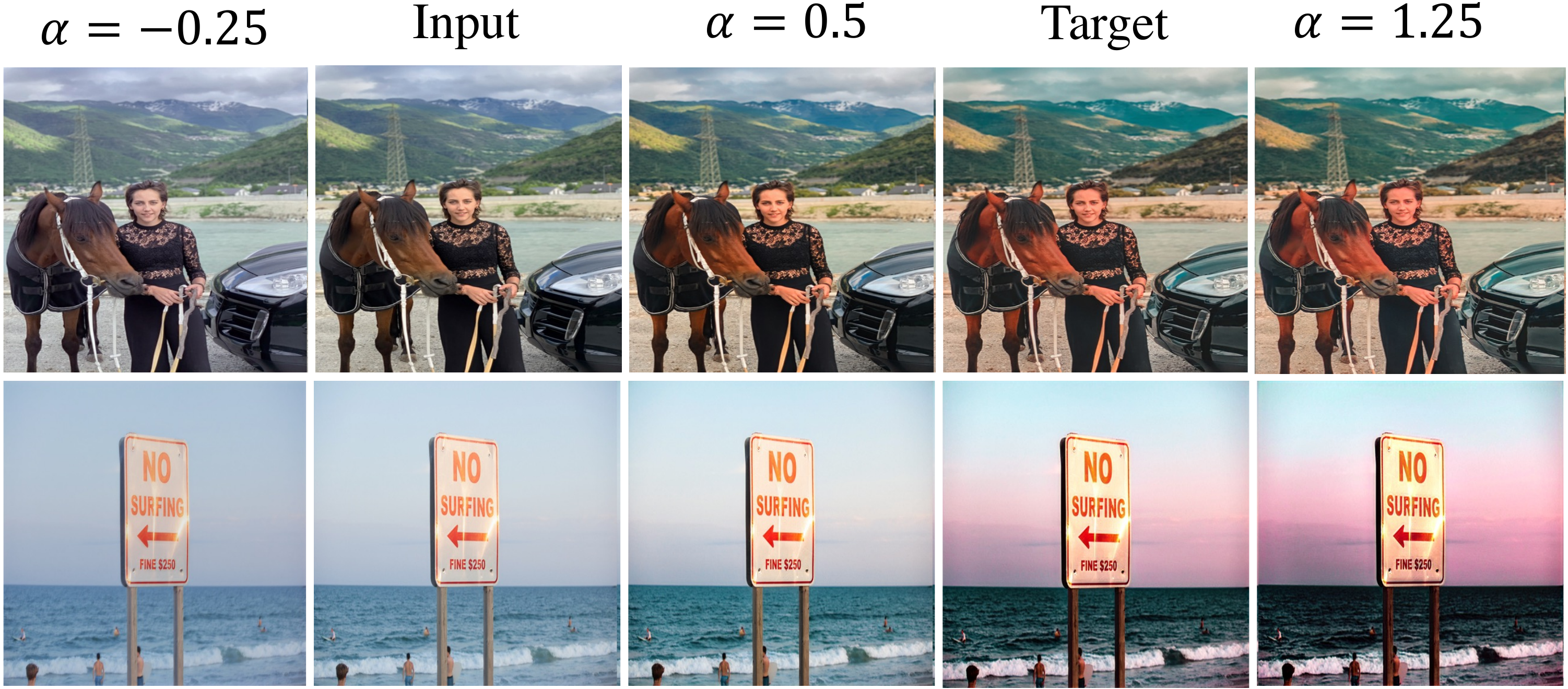}
    \caption{From the left to right, the strength the editing style increases.}
    \label{fig:interpolation_cluster}
\vspace{-6mm}
\end{figure}

\subsection{Interpolation}
A special case of the conditional GAN inversion, which has not been investigated in the previous literature, is to find a latent code $\mathbf w_0$ that can reconstruct the source image itself. Such latent code has some semantic meaning in terms of editing as it represents the unchanged status of the source image. We can find its embedding by simply replacing the $I_{tgt}$ term with $I_{in}$ in Eq.~(\ref{eqn:condGAN_inv}).
The reconstruction error on the testing dataset is less than 2 pixel difference as shown in Tab.~\ref{tab:condGAN_inv}.

With the help of $\mathbf w_0$, we can control the strength of an arbitrary editing style $\mathbf w$ by using their linear interpolations as $\mathbf w' = (1 - \alpha)\mathbf w_0 + \alpha\mathbf w$, where $\alpha$ is a factor to control the strength of editing. Some examples are shown in Fig.~\ref{fig:interpolation_cluster}. 

\subsection{Other Properties}
\label{sec:property}
We further demonstrate the editing capability and recognition capability of $\mathcal{W}$ space.
For editing capability, as Fig.~\ref{fig:viz_style} reveals that each $\mathbf w$ shows a consistent style for different images, enabling the transfer of $\mathbf{w}$ inverted from one image pair to other images to achieve similar editing style, indicating its \emph{transferability} property as shown in Fig.~\ref{fig:viz_transfer}, detailed in Sec.~\ref{sec:transfer}.
For recognition capability, we demonstrate that the latent codes representing similar editing styles are distributed closely in $\mathcal{W}$ space by studying the retrieval and cluster performance in $\mathcal{W}$ space (see Sec.~\ref{sec:disentangle}), showing that the latent code has the intrinsic capability to be used for recognize the editing style.

\section{Language-Guided Image Editing}
To show the advantage of our pretrained network on the downstream tasks, we firstly show the language-guided image editing~(LGIE) by leveraging our pretrained model. Other downstream tasks are illustrate in Sec.~\ref{sec:exp_downstream}.
Given an image $I$, and a language editing request $r$, LGIE aims to generate a new image following the editing request.
Language is a convenient way to incorporate user's editing intention, which is a more intuitive and convenient interface than existing operation-based editing interfaces. 
Given our pretrained generator, we solve the LGIE tasks by finding a mapping between the text input and our low-dimensional editing space, which is a different framework compared to previous works~\cite{chen2018language,el2019tell,shi2020benchmark,shi2021learning,jiang2021language,dong2017semantic,nam2018text,mao2019bilinear,li2020manigan,yu2019multi,bau2021paint,liu2020open}. Next, we describe our approaches for both supervised LGIE as well as zero-shot LGIE.

\begin{figure}[t]
	\centering
	\includegraphics[width=1\linewidth]{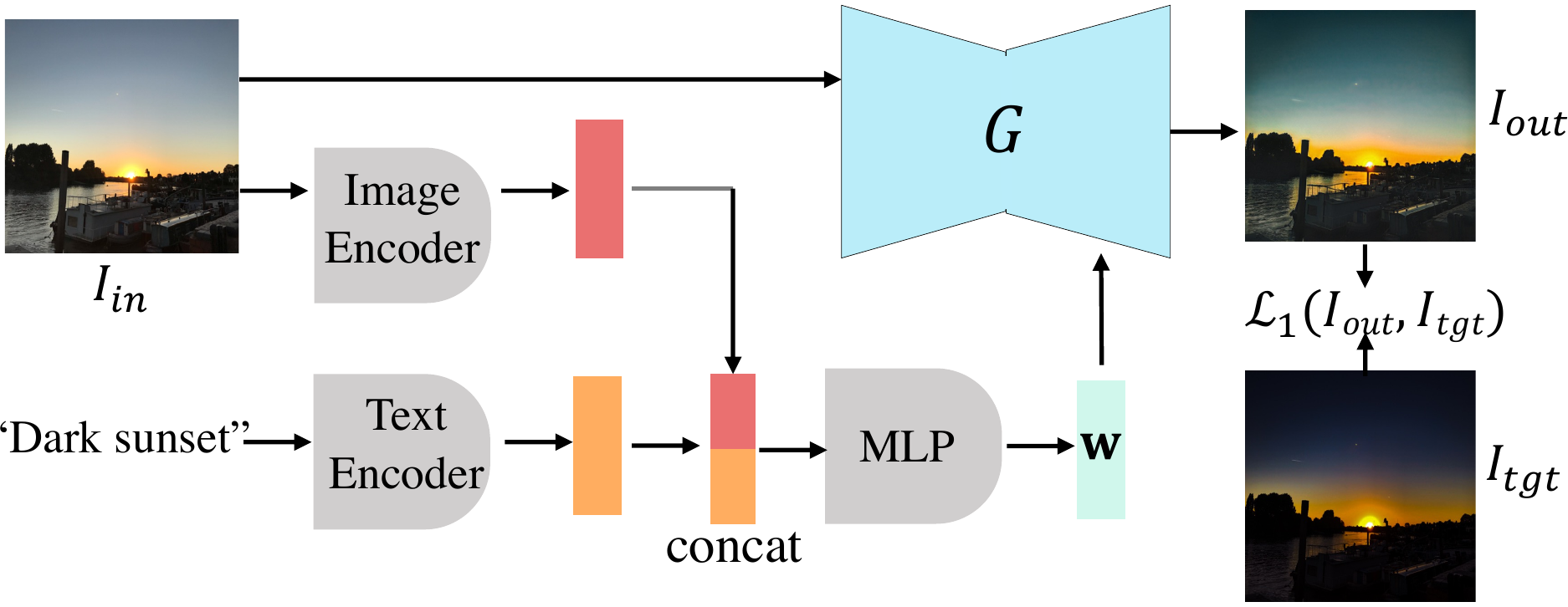}
    \caption{The structure for supervised LGIE. Only gray shaded module are trained while the generator is frozen.}
    \label{fig:lgie_structure}
\vspace{-6mm}
\end{figure}

\noindent\textbf{Supervised LGIE.}
The supervised LGIE directly learns the mapping from language to the $\mathcal{W}$ space from the data triplet consisting of the input image, target image, and language request. 
The structure of the model is shown in Fig.~\ref{fig:lgie_structure}, where the image and text feature are merged by concatenation, followed by a Multilayer Perception~(MLP) to predict a latent code $\textbf w$. Given $\textbf w$, the generator serves as a render to generate the output image with the designated style.
The training is driven by the L1 loss between the output image and target image, written as $\mathcal{L}_1(I_{out}, I_{tgt})$.
The generator $G$ is frozen while the other parameters are trained.
Our novel learning framework could be potentially useful for other image editing tasks with paired supervision, such as supervised image harmonization, which will be left for future study.

\noindent\textbf{Zero-Shot LGIE.}
Inspired by StyleCLIP~\cite{patashnik2021styleclip}, we propose to use the pretrained image-text CLIP model~\cite{radford2021learning} to directly find a latent code $\mathbf w$ given an editing request $r$ through optimization.
Specifically, given the CLIP visual encoder $f_{v}$ and textual encoder $f_{t}$, the latent code $\mathbf{w}$ is optimized by
\begin{equation} \label{eqn:clip}
    \arg\min_{\mathbf{w}} -\left\langle f_{v}(G(I, \mathbf{w})), f_{t}(r)\right\rangle + \lambda \left\langle f_{v}(G(I, \mathbf{w})), f_{v}(I) \right\rangle, 
\end{equation} 
where $\left\langle \cdot, \cdot\right\rangle$ denotes the cosine similarity and $\lambda$ a balance weight. 
Its first term enforces the CLIP similarity between the generated image and the request. The second term drives the similarity of the generated image to the original image.
Since the CLIP model is trained on billions of image-text pairs and thus understands free-form language, this approach is generic for open-vocabulary requests.

Moreover, to achieve precise local editing, our approach can accepts as input an additional binary mask $M$ to indicate the editing foreground and background. Given an editing request, we can simply replace the term $G(I,\mathbf{w})$ in Eq.~(\ref{eqn:clip}) with $M\odot G(I, \mathbf{w})  + (1 - M) \odot I$, where $\odot$ is Hadamard product.

\begin{figure}[t]
	\centering
	\includegraphics[width=1\linewidth]{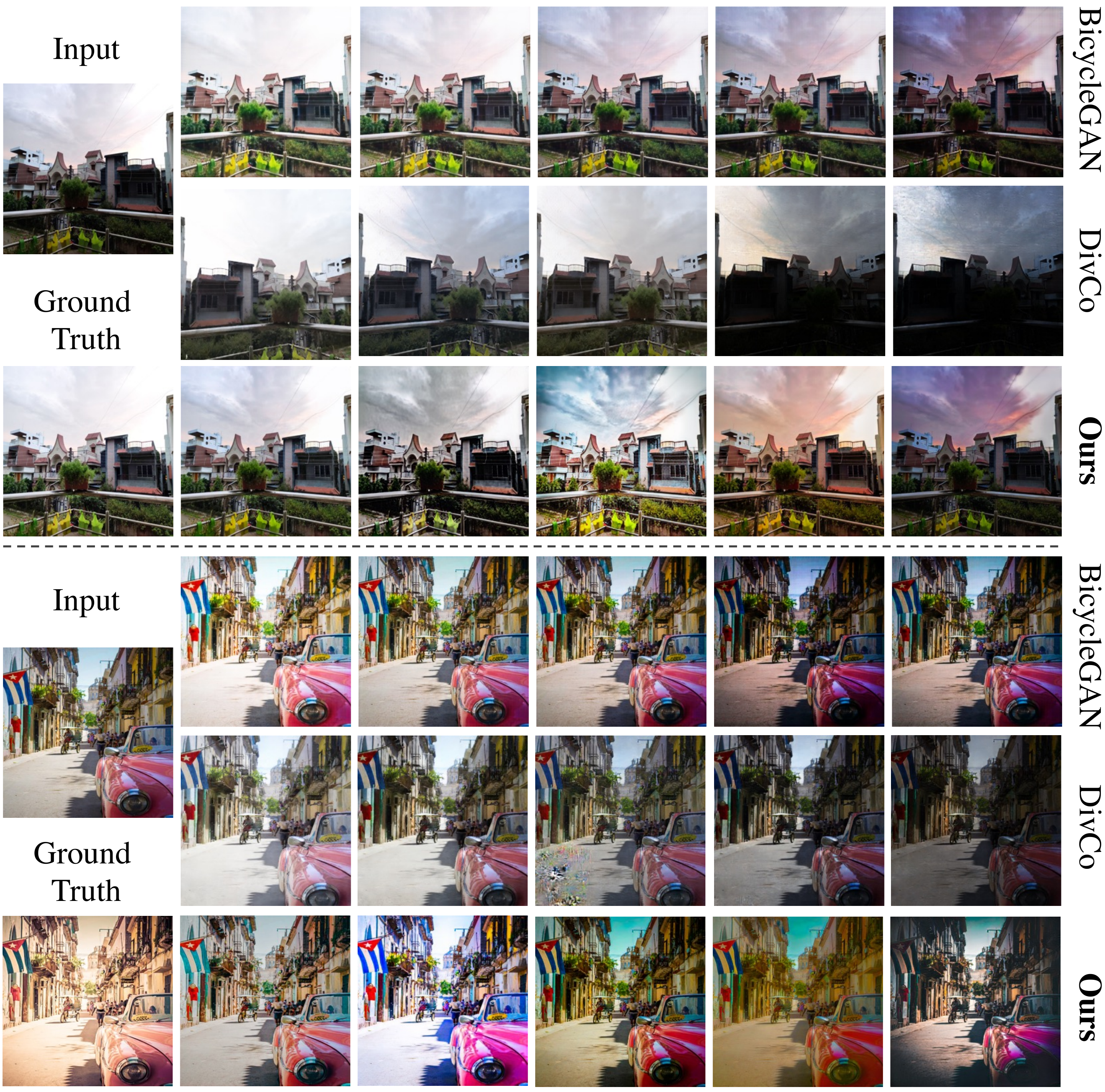}
    \caption{The multimodal image editing performance compared with other methods}
    \label{fig:viz_retouching}
\vspace{-5mm}
\end{figure}

\section{Experiments} \label{sec:expriment}
We evaluate the pretraining task, $\mathcal{W}$ properties, and downstream tasks in this section. Due to space limitation, we put the implementation details in Appx.~\ref{appx:implement_detail}.

\subsection{Multimodal Image Editing}

\noindent\textbf{Dataset.} 
We use the Adobe Discover dataset collected from the Adobe Discover website, where Lightroom users upload their edited images along with editing operations. This paired dataset contains open-domain images with various editing styles, focusing on color and tone retouching while not changing image content, geometry, or texture.
Given the large number of active users, totally 62416 before- and after-image pairs are collected with the split of 49932/6242/6242 for train/val/test.

\noindent\textbf{Metrics.}
\textit{Fr\'echet Inception Distance}~(FID)~\cite{heusel2017gans} measures the quality and diversity of a set of generated images compared to the set of real images through the feature computed from an Inception network~\cite{szegedy2015going}. 
A lower value indicates higher visual quality and diversity. 
LPIPS~\cite{zhang2018perceptual} measures the diversity of an image set by computing the average feature distance of all pairs of images, following the setting of \cite{zhu2016generative}. For each input, we generate 10 random outputs for computing LPIPS.
A higher value denotes higher diversity.

\noindent\textbf{Comparison methods.}
\noindent\textit{BiCycleGAN}~\cite{zhu2017multimodal} learns the mapping from the output image to the input noise to encourage diversity.
\noindent \textit{DivCo}~\cite{liu2021divco} follows the structure of BiCycleGAN but adds the contrastive loss to encourage better diversity.

\noindent\textbf{Result analysis.}
Our algorithm surpasses BiCycleGAN and DivCo by a large margin according to FID, mainly due to the benefit of the styleGAN-like structure.
And as indicated in \cite{zhao2021large}, the modulation-based conditional generator is intrinsically stochastic w.r.t. the input noise even without explicit diversity constraint used in ~\cite{zhu2017multimodal,liu2021divco}.
The qualitative comparison in Fig.~\ref{fig:viz_retouching} shows that our model can create more diversified editing styles, while the BicycleGAN and DivCo will only generate images in a single editing style with different degrees.
Moreover, we sample the same $\mathbf{z}$ for different images in Fig.~\ref{fig:viz_style}, showing that the same $\mathbf{z}$ ($\mathbf{w}$) has global consistency for all the images.

\noindent\textbf{Ablation Study of the network structure.}
Firstly, since the study of Sec.~\ref{sec:disentangle} suggests that our editing space takes most effect at high-resolution layers of the decoder, we remove the deeper layers of both encoder and decoder and only keep the layers sensitive to $\textbf{w}$, so as to reduce the model size.
We denote such setting as \emph{Ours shallow}, whose performance in Tab.~\ref{tab:multi-modal-exp} is worse than the standard setting.
Therefore, it proves that the depth of the network is still critical for editing performance.

Moreover, our standard network is only modulated by the noise input, while it also can be co-modulated by the feature extracted from the input image, similar to the structure of \cite{zhao2021large}.
We therefore compare this setting as \emph{Ours comod} in Tab.~\ref{tab:multi-modal-exp}.
However, the performance for co-modulation drops. One possible reason is that the image modulation features bring some input-constrained information which impairs the editing quality and stochasticity.

\begin{table}[t]\centering
\ra{1}
\scalebox{0.8}{
\begin{tabular}{@{}lrr@{}}
\toprule
& FID$\downarrow$ & LPIPS $\uparrow$\\
\midrule
BiCycleGAN~\cite{zhu2017multimodal}  & 12.2837 & 0.0857 \\
DivCo~\cite{liu2021divco} & 9.9586 & 0.1705 \\
Ours & \textbf{5.1755} & \textbf{0.1945}  \\
\midrule
Ours shallow & 6.0958 & 0.1581 \\
Ours comod~\cite{zhao2021large} & 5.6355 & 0.1479  \\
\bottomrule
\end{tabular}}
\caption{Quantitative results of multimodal image editing on Discover dataset.}
\label{tab:multi-modal-exp}
\vspace{-6mm}
\end{table}

\begin{figure}[t]
	\centering
	\includegraphics[width=1\linewidth]{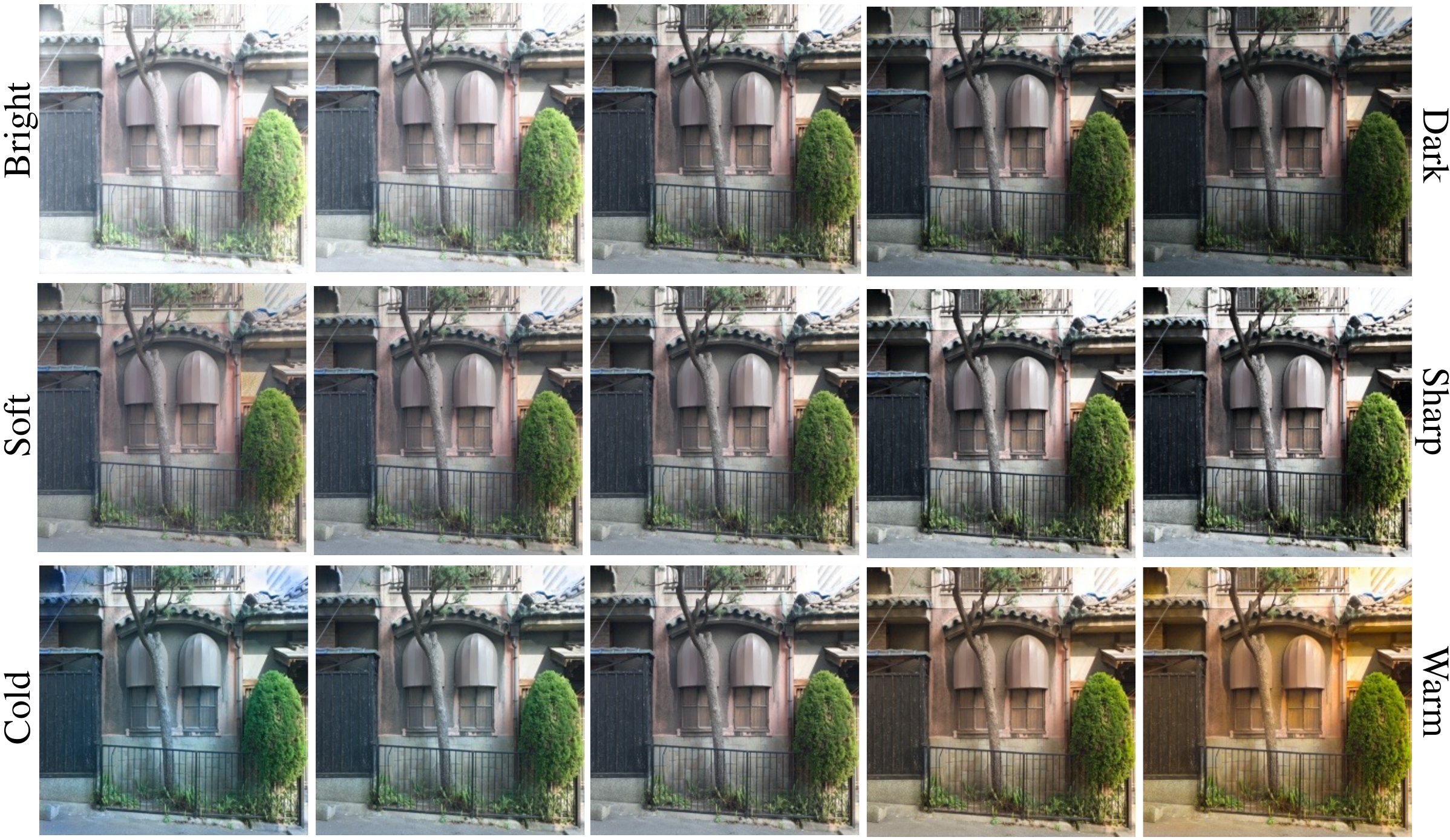}
    \caption{The visualization for unsupervised latent direction discovery using SeFa. The center column is the input image, and each row is the traverse through one SeFa principle direction across $w_0$. }
    \label{fig:sefa}
\vspace{-2mm}
\end{figure}

\subsection{Latent Space Analysis}
\label{sec:disentangle}
We analyze the semantics of the editing space $\mathcal{W}$ with the following experiments.

\noindent\textbf{Semantic disentanglement.}
Given the line of works~\cite{voynov2020unsupervised,collins2020editing,wang2021geometry,shen2021closed,wu2021stylespace} tackling unsupervised GAN latent semantic discovery, we adopt \emph{Semantic Factorization} (SeFa)~\cite{shen2021closed} for the sake of simplicity.
Some discovered principal semantic direction is visualized in Fig~\ref{fig:sefa}, showing that the editing space $\mathcal{W}$ can be disentangled.

\noindent\textbf{Layerwise effect of $\mathbf{w}$}.
Similar to StyleGAN, our $\mathbf{w}$ applies to different layers of the decoder.
So we further analyze its layerwise semantics using SeFa.
We find that the editing is only caused by the $\mathbf{w}$ on high-resolution layers, while the effect of $\mathbf{w}$ in low-resolution layers is not obvious.
Concretely, $\mathbf{w}$ is most effective for the top 6 out of 14 layers in the decoder for 256\textsf{x}256 resolution input.
This is reasonable since our model focus on color manipulation which is typically controlled via the top layers of the styleGAN~\cite{yang2021semantic}.
However, we cannot tell obvious semantic differences among the top layers, as shown in Appx.~\ref{appx:effect_w}, which might be because the color adjustment is already located in a fine-grained subspace.

\begin{figure}[t]
	\centering
	\includegraphics[width=1\linewidth]{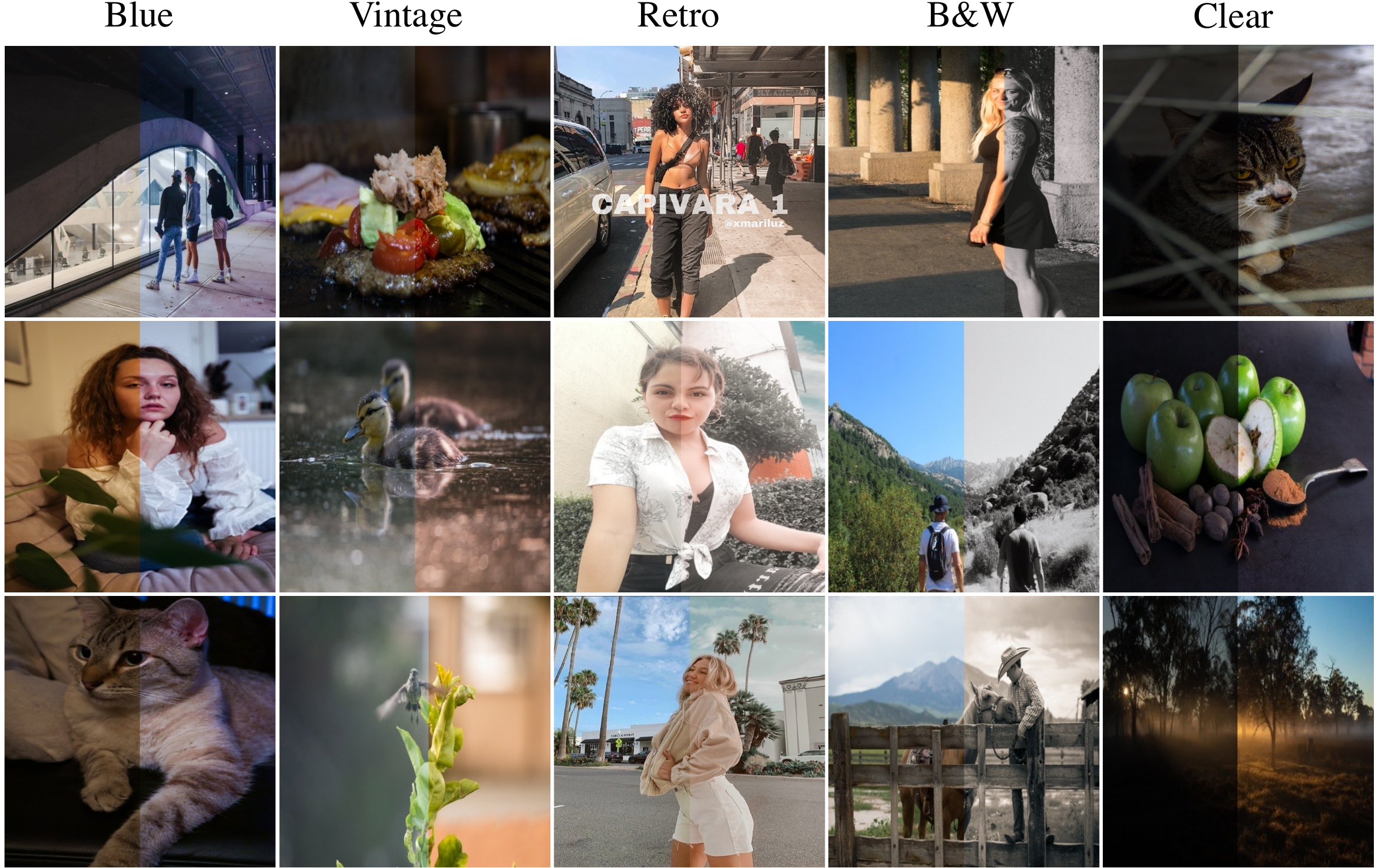}
    \caption{The clustering of the dataset using $w$. For each image, the left half is the before-image, and the right half is the after-image.}
    \label{fig:viz_cluster}
\vspace{-6mm}
\end{figure}

\begin{table}[t]\centering
\ra{1}
\scalebox{0.8}{
\begin{tabular}{@{}ccc|c@{}}
\toprule
 & Lr Operation & $\mathcal W$ (ours) & $\mathcal W$ (euc) \\
\midrule
Purity $\uparrow$ & 4.25 & \textbf{12.76} & 11.30 \\
\bottomrule
\end{tabular}}
\caption{Quantitative clustering results on Discover dataset. Euc denotes cluster using euclidean distance.}
\label{tab:cluster_purity}
\vspace{-3mm}
\end{table}

\noindent\textbf{Retrieval capability.}
Next, we assess the distribution of different editing styles in the editing space $\mathcal{W}$.
We conduct k-nearest neighbor~(KNN) search in the database using inverted $\mathbf{w}$ with cosine distance. 
Given a pair of before- and after-images as query, the retrieved KNN image pairs carry the similar editing style, shown in Fig.~\ref{fig:teasing} (see more in Appx.~\ref{appx:retrieval})
The retrieval result illustrates that the similarity in the $\mathcal{W}$ space measures the similarity of editing style.

\noindent\textbf{Clustering capability}. Inspired by the retrieval result, it uncovers another simple way for latent semantic discovery -- cluster in the $\mathcal{W}$ space and regard each cluster center as an editing style.
We employ K-means algorithm with cosine distance for clustering.
To evaluate the cluster performance, ideally we need to annotate the style class for each editing pair. 
However, as the editing styles in the dataset are diversified and compositional, a predefined list of style tags might be short-sighted.
So we instead annotate a complete sentence that describes the edit, allowing novel styles to be included.
Then we create a style tag list including both common styles and the novel styles mentioned in the labeled sentences.
Next, we evaluate the clustering performance by purity which is a measure of the extent to which clusters contain a single class. 
As the standard purity only considers the data sample with the single-class label, while our sample (image pair) bears multiple style tags. 
Hence we customize the computation of purity in Appx.~\ref{appx:purity}.

For comparison, as the Adobe discover dataset also contains the ground-truth Lr operation parameter, we compare our editing space with the Lr operation space. 
The result shown in Tab.~\ref{tab:cluster_purity} indicates that our editing space has better semantics to represent styles than the Lr operation space. 
Moreover, we compare the default cosine distance with the euclidean distance and find that the cosine distance is better.
Fig.~\ref{fig:viz_cluster} shows the representative tag for some clusters.
Due to space limit, the details for the tag list and annotation process are in Appx.~\ref{appx:tag_list} and \ref{appx:data_collection}.

\begin{table}[t]\centering
\ra{1}
\scalebox{0.8}{
\begin{tabular}{@{}lrrrr@{}}
\toprule
\cmidrule{2-5} 
& L1 $\downarrow$ & SSIM$\uparrow$ & FID$\downarrow$ & $\sigma_{\times 10^2}$ $\uparrow$\\
\midrule
Input & 0.1190 & 0.7992 & 12.3714 & - \\
T2ONet~\cite{shi2021learning} & 0.0784 & 0.8459 & 6.7571 & \textbf{0.7190} \\
EDNet~\cite{jiang2021language} & - & - & 9.9500 & -  \\
Ours & \textbf{0.0731} & \textbf{0.8721} & \textbf{5.9791} & 0.6809  \\
\midrule
Ours w/o vis & 0.0795 & 0.8596 & 6.9757 & 0.6281 \\
\bottomrule
\end{tabular}}
\caption{Quantitative results on MA5k-Req test sets. $\sigma_{\times 10^2}$ denotes the image variance scaled by 100 times.}
\label{tab:LGIEResult}
\vspace{-6mm}
\end{table}

\begin{figure}[t]
	\centering
	\includegraphics[width=1\linewidth]{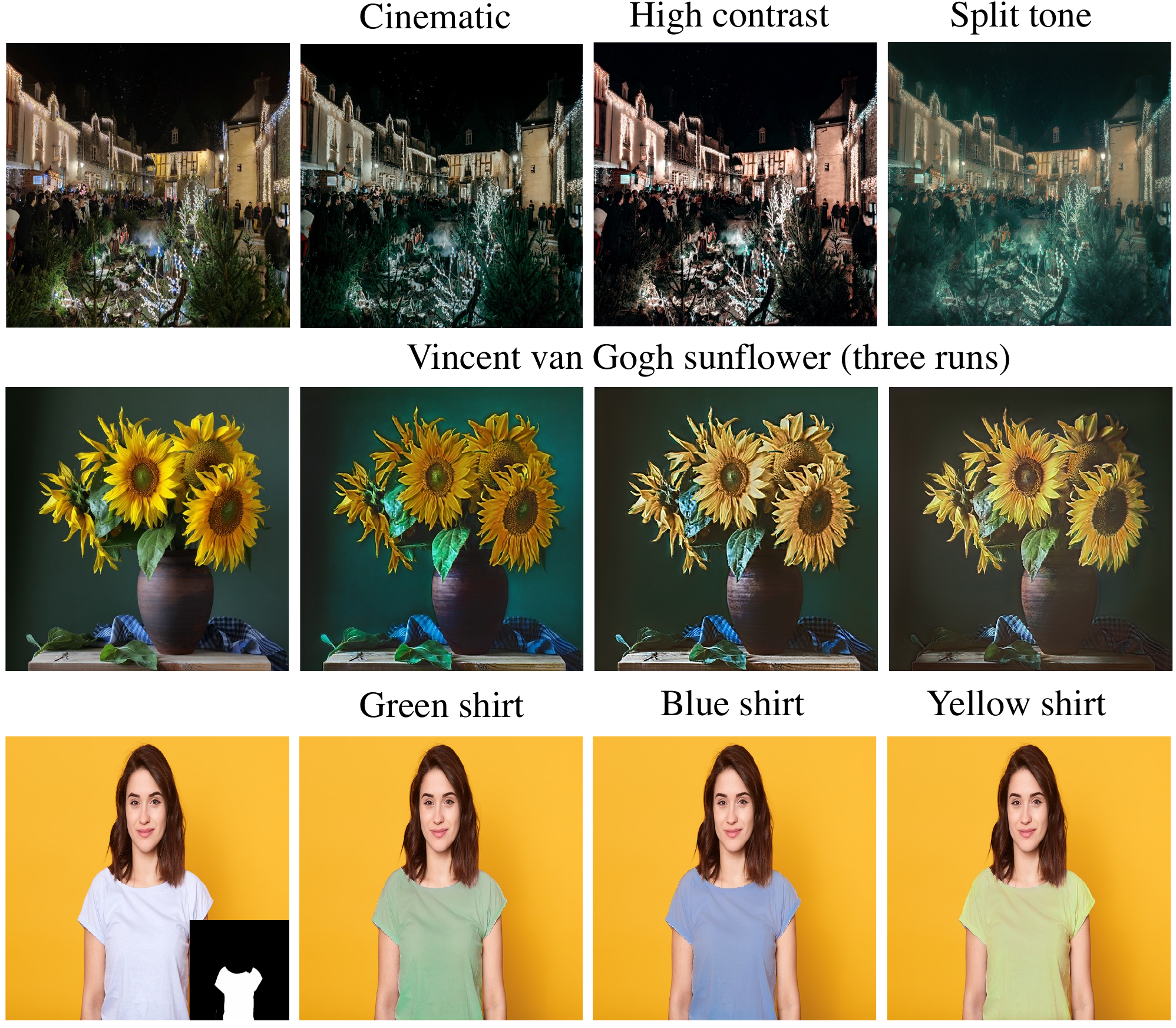}
    \caption{The open-vocabulary, open-image, language-guided image editing samples optimized by CLIP. The last row show the local editing with mask input.}
    \label{fig:viz_clip}
\vspace{-2mm}
\end{figure}

\subsection{Downstream Tasks}
\label{sec:exp_downstream}

\subsubsection{Language-guided image editing}
\textbf{Experimental settings.} For supervised LGIE, we follow the experiment setting of the previous work~\cite{shi2021learning} on the MA5K-Req~\cite{shi2021learning} dataset, which is an extension of the MIT-Adobe FiveK dataset~\cite{bychkovsky2011learning} with additional language augmentation. 
For metrics, we evaluate L1, SSIM, FID, and image variance $\sigma$. Due to the space limit, we put the detailed description in Appx~\ref{appx:supervised_LGIE}.
We compare our method with two SOTA methods \emph{T2ONet}~\cite{shi2021learning} and \emph{EDNet}~\cite{jiang2021language}, as well as a base evaluation between the input and output images denoted as \emph{Input}.

For zero-shot LGIE, as it works for open-domain image and open-vocabulary requests, we compare the qualitative performance on given examples with two other SOTA methods -- OpenEdit~\cite{liu2020open} and StyleCLIP~\cite{patashnik2021styleclip}. OpenEdit has no constraint for both image and request, while StyleCLIP can only work for close-domain images.

\noindent\textbf{Result analysis.}
For the supervised LGIE, the performance is shown in Tab.~\ref{tab:LGIEResult}, showing that our method achieves the best editing quality and comparable variance as T2ONet, demonstrating the advantage of the pretrained generator.
Given the strong editing ability of the pretrained generator, the LGIE task becomes easier because the model only needs to predict a latent code of 512 dimensions instead of the entire image space.
Moreover, we study whether the language input alone is sufficient to predict the latent code.
We denote the setting without image input as \textit{ours w/o viz} shown in Tab.~\ref{tab:LGIEResult}, which shows inferior results to the standard setting, thus suggesting the importance of the visual input.

\begin{figure}[t]
	\centering
	\includegraphics[width=1\linewidth]{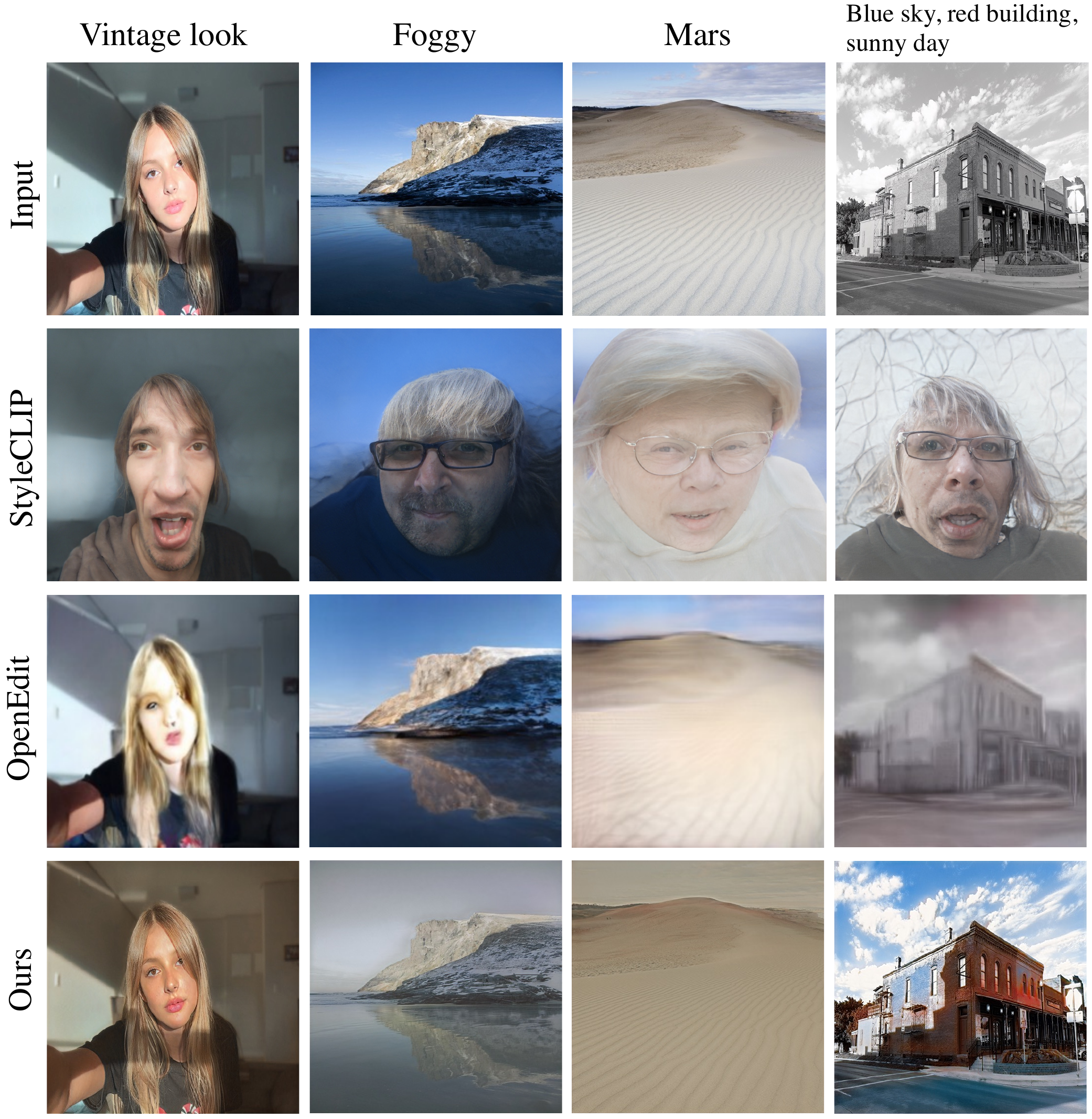}
    \caption{The open-vocabulary, open-image, language-guided image editing samples optimized by CLIP with comparison to other methods.}
    \label{fig:viz_clip_comp}
\vspace{-3mm}
\end{figure}

For the zero-shot LGIE, we firstly show our result in Fig.~\ref{fig:teasing} and~\ref{fig:viz_clip}, indicating that our model can achieve the editing with the diversified directive of high-level semantic (aurora), editing terminology (split tone), color manipulation (green shirt), or even some texture change (Van Gogh painting). 
Furthermore, the comparison with the SOTA is drawn in Fig.~\ref{fig:viz_clip_comp}. 
StyleCLIP completely fails in these cases because it does not work for open-domain images. The face will pop up due to the memory of its generator pretrained on face dataset. 
Despite that OpenEdit can accept open-domain images, its editing does not follow the request, and the output image contains obvious artifacts.
In contrast, our method can handle these cases well.
Despite imperfect, our model has the potential to achieve gray image colorization while other methods cannot.
\vspace{-3mm}

\subsubsection{Personalized Editing and Recommendation}
\label{sec:transfer}
\begin{figure}[t]
	\centering
	\includegraphics[width=1\linewidth]{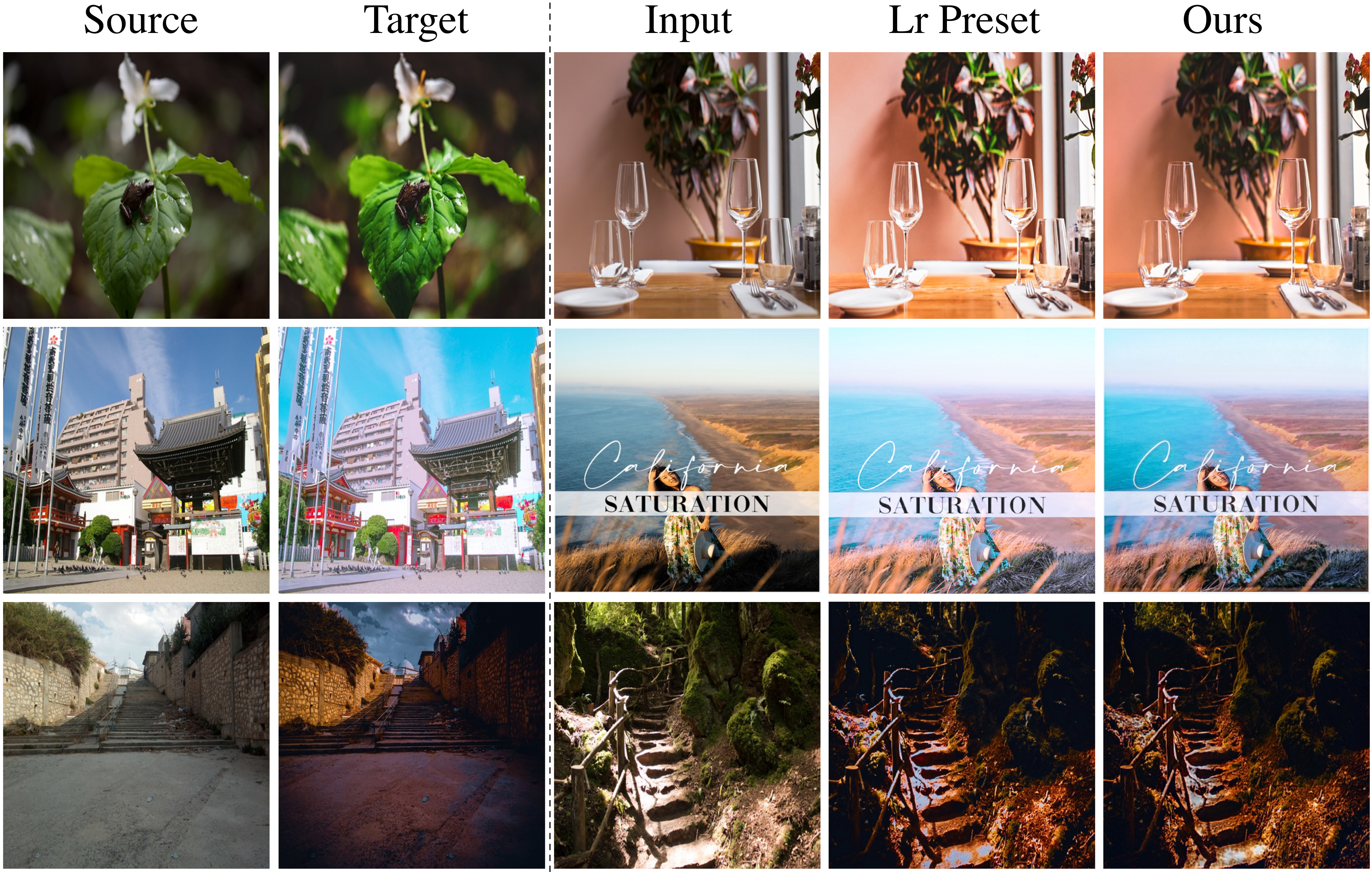}
    \caption{The visualization of the exemplar-based image editing. The left of the dash line are exemplars and the right is the transferred editing.}
    \label{fig:viz_transfer}
\vspace{-3mm}
\end{figure}

Given a user-edited before- and after-image pair as an exemplar, our model can achieve both personalized editing and editing style recommendation.
For personalized editing, we study exemplar-based image editing~(EBIE), which is to edit the input image following the editing style of the exemplar. 
It helps personalized editing because it can propagate the user's preferred editing style to new images.
This task can be naturally tackled by the transferability property~(Sec.~\ref{sec:property}) of the $\mathcal{W}$ space without training.
When there are multiple exemplars with consistent styles, we can find a common editing direction by averaging the latent code of all the exemplars.
We compare our approach with the Lr preset, which can also be applied to other images to achieve a similar editing effect.
The visualization of the comparison is shown in Fig.~\ref{fig:viz_transfer}, indicating our transfer result is reasonable and visually comparable with the Lightroom preset. 
However, the preset approach must know the exact preset parameters of the exemplar images, while our method is free from such constraint and thus is more general.
Moreover, different from the photorealistic style transfer~\cite{yoo2019photorealistic} where the color and texture of the reference image is directly transferred to the source image, our EBIE tries to transfer the relative editing style.
Taking the first row of Fig.~\ref{fig:viz_transfer} as an example, our method transfer the ``brighten" effect instead of the green color to the other image.

In addition, we can achieve editing recommendation, which is to recommend the image pair with similar editing styles to a given image pair.
This task is beneficial for the photography pedagogy if a user wants to see multiple photo examples of the same editing style for specialized learning.
Such task can be handled via the retrieval capability in the $\mathcal{W}$ space, as illustrated in Sec.~\ref{sec:disentangle}.
The visualization is shown in Appx.~\ref{appx:retrieval}.

\section{Conclusion and Discussion}
\label{sec:conclusion}
This paper introduces a new image editing paradigm: learn a pretrained I2I generator with an editing space that can work as a unified interface to bridge multiple downstream tasks.
We find the editing space is well disentangled and complete for color editing, which can be used for both editing and recognition. 
Experiments on the downstream tasks prove the advantages of our pretrained model.

\noindent\textbf{Limitation.} Our method relies on the Adobe Discover dataset and thus cannot be expected to manipulate image content (\eg geometric change) or texture (though we have shown some particular texture changes in painting style, they are not general).
For LGIE, a faithful image manipulation is not guaranteed if the text requests are mapped to the CLIP space where images are not well populated.

\noindent\textbf{Potential Negative Impact.}
Our model might be maliciously used to generate fake photos to forge criminal evidence, \eg, daytime to night. Therefore we keep the user's identity and editing history to monitor misuse.

{\small
\bibliographystyle{ieee_fullname}
\bibliography{egbib}
}
\clearpage
\noindent{{\huge \textbf{Appendix}}}
\appendix

\section{Network Structure}\label{appx:network_structure}
\begin{figure}[t]
	\centering
	\includegraphics[width=1\linewidth]{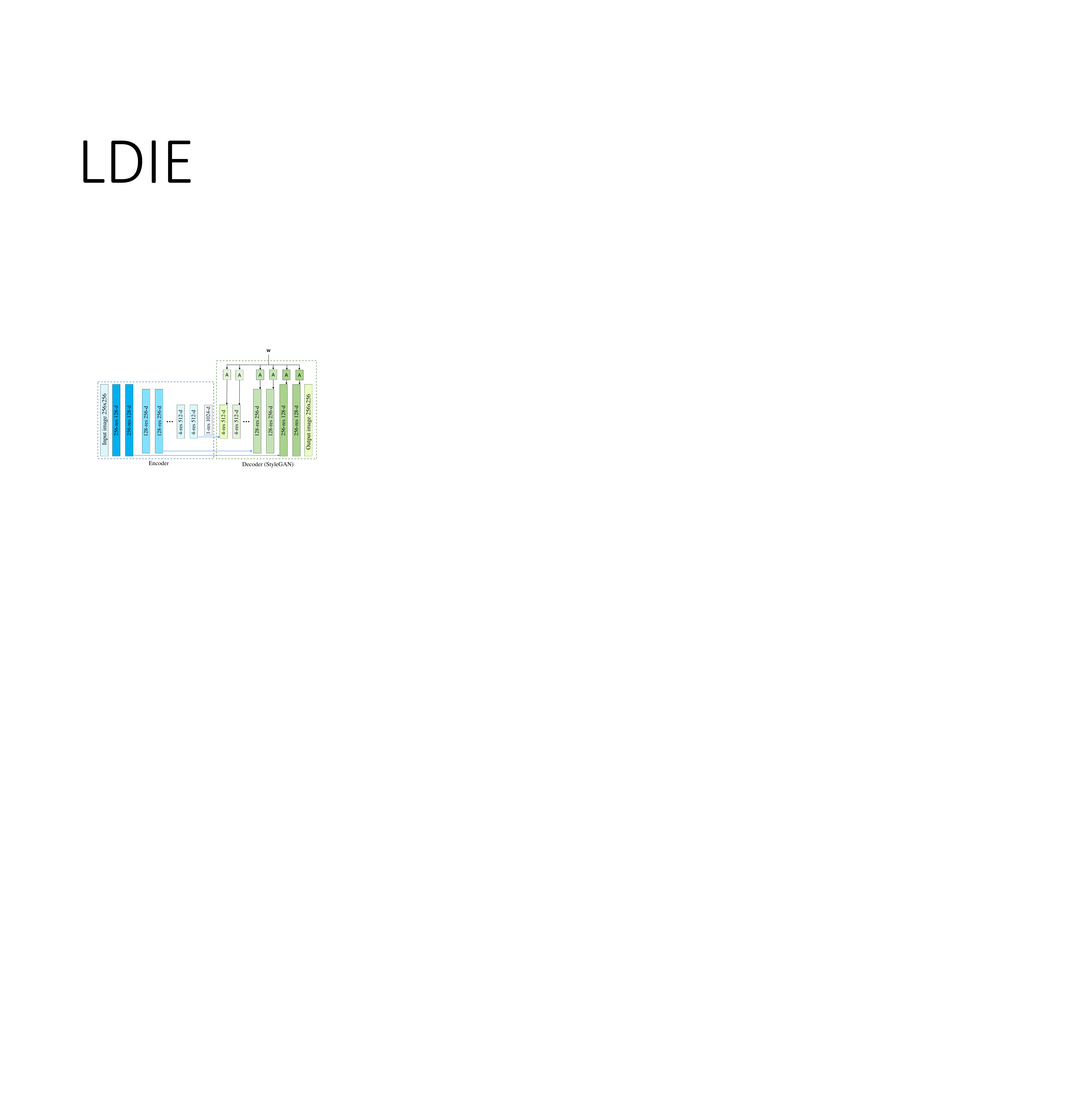}
    \caption{The details of our encoder and the skip connection.}
    \label{fig:appx_network_structure}
\vspace{-2mm}
\end{figure}

We take the structure of the generator for 256x256 image as an example and illustrate the model structure in detail in Fig.~\ref{fig:appx_network_structure}.
Note that the mapping from $\mathbf{z}$ to $\mathbf{w}$ are an MLP which is the same as StyleGAN2~\cite{karras2020analyzing}, and both $\mathbf{z}$ and $\mathbf{w}$ are 512-dimensional.
The whole structure of the decoder is the same as StyleGAN2 except that the skip connection is added, which is denoted as blue arrow.
Since the $\mathbf{w}$ is operated actually at the conv-layer after each feature map, so the position of the arrow from $\mathbf{w}$ is shown at the end of each feature map in the decoder.
The skip connection will add the feature of the encoder to the corresponding feature of the decoder.
The encoder is basically the inverse structure of the decoder, and the feature size and dimension are shown in Fig.~\ref{fig:appx_network_structure}, which are obtained through convolutional layers.
The omitted layers in the encoder follow the same rule that the resolution will reduce twice and the dimension will increase twice at maximum 512 dimension every two conv-layers.

\section{Implementation Details} \label{appx:implement_detail}
The whole project is implemented with Pytorch~\cite{paszke2019pytorch}. 
\subsection{Multimodal Image Editing}
The model for visualization and the comparison methods are for 256x256 images.
The input image is normalized to $(-1, 1)$
The model is optimized by Adam~\cite{kingma2014adam} with learning rate 0.0025, $\beta_1=0$, $\beta_2=0.99$. 
We totally train the model for 5 million images, with batch size 32 on 8 V100 GPUs for 1 day.
We also trained the generator for 512x512 images for language-guided image editing.

\subsection{Language-Guided Image Editing}
We use the pretrained generator for 512x512 images.

\textbf{Supervised LGIE.} 
The image encoder is ResNet50~\cite{he2016deep}, text encoder is the text transformer from the CLIP model~\cite{radford2021learning}.
The image resolution is 512x512 and the input image is normalized to $(0, 1)$
The model is trained with Adam~\cite{kingma2014adam} with learning rate 0.0001, $\beta_1=0.9$, $\beta_2=0.999$.

\textbf{Zero-shot LGIE.}
The editing process is optimized with the same optimizer and hyperparameter as the GAN inversion process in StyleGAN2~\cite{karras2020analyzing}.
Moreover, the balance weight $\lambda$ is flexible. 
We will output all the edited results given different $\lambda$ ranging from 0.1 to 0.5 and then we select the best one.

\subsection{Retrieval and Clustering}
We need to conduct the conditional GAN inversion for all the dataset to obtain the $\mathbf{w}$ to support the style retrieval and clustering.
Therefore, to accelerate the speed, we inverse the $\mathbf{w}$ corresponding to 128x128 resolutional images. 
We follow the same training setting for the inversion as StyleGAN2~\cite{karras2020analyzing}.
The average time for such inversion is 30s per sample.
Then we use KNN with cosine distance on the $\mathcal{W}$ space for retrieval and k-means with cosine distance on the $\mathcal{W}$ for clustering.
Even though the $\mathbf{w}$ is only for 128x128 generator, it will not harm the output performance because in this stage we do not use $\mathbf{w}$ to generate images.

\subsection{Examplar-Based Image Editing}
We conduct the conditional GAN inversion for 512x512 resolutional generator, and transfer the inverted $\mathbf{w}$ to new 512x512 resolutional images.

\section{Effect of w at Different Layers} \label{appx:effect_w}
\begin{figure*}[t]
	\centering
	\includegraphics[width=1\linewidth]{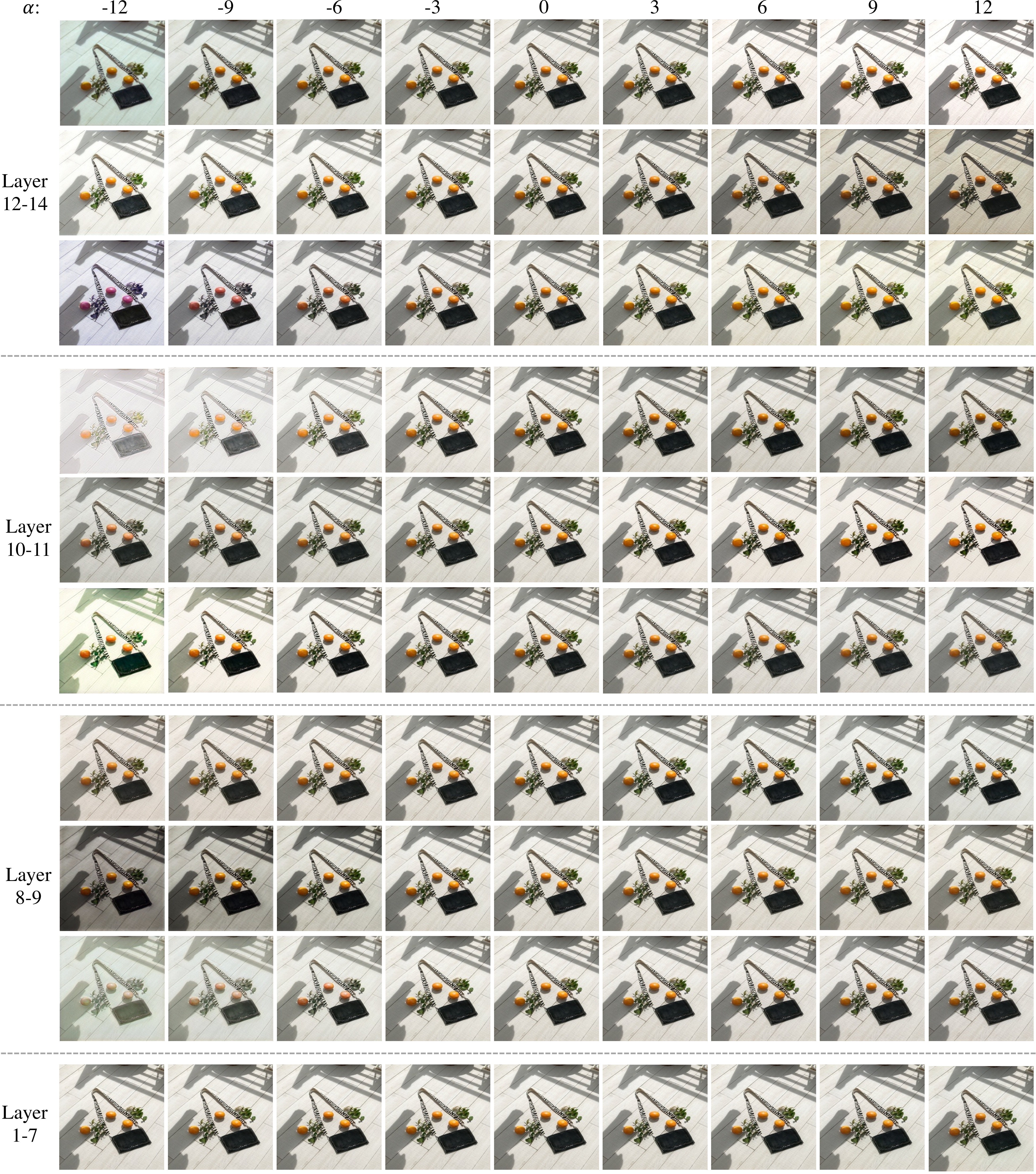}
    \caption{The visualization of the SeFa disentanglement on different layers. We select top-3 principle directions in layer 12-14, 10-11, 8-9 and top-1 direction for layer 1-7.}
    \label{fig:layerwise_sefa1}
\vspace{-2mm}
\end{figure*}

\begin{figure*}[t]
	\centering
	\includegraphics[width=1\linewidth]{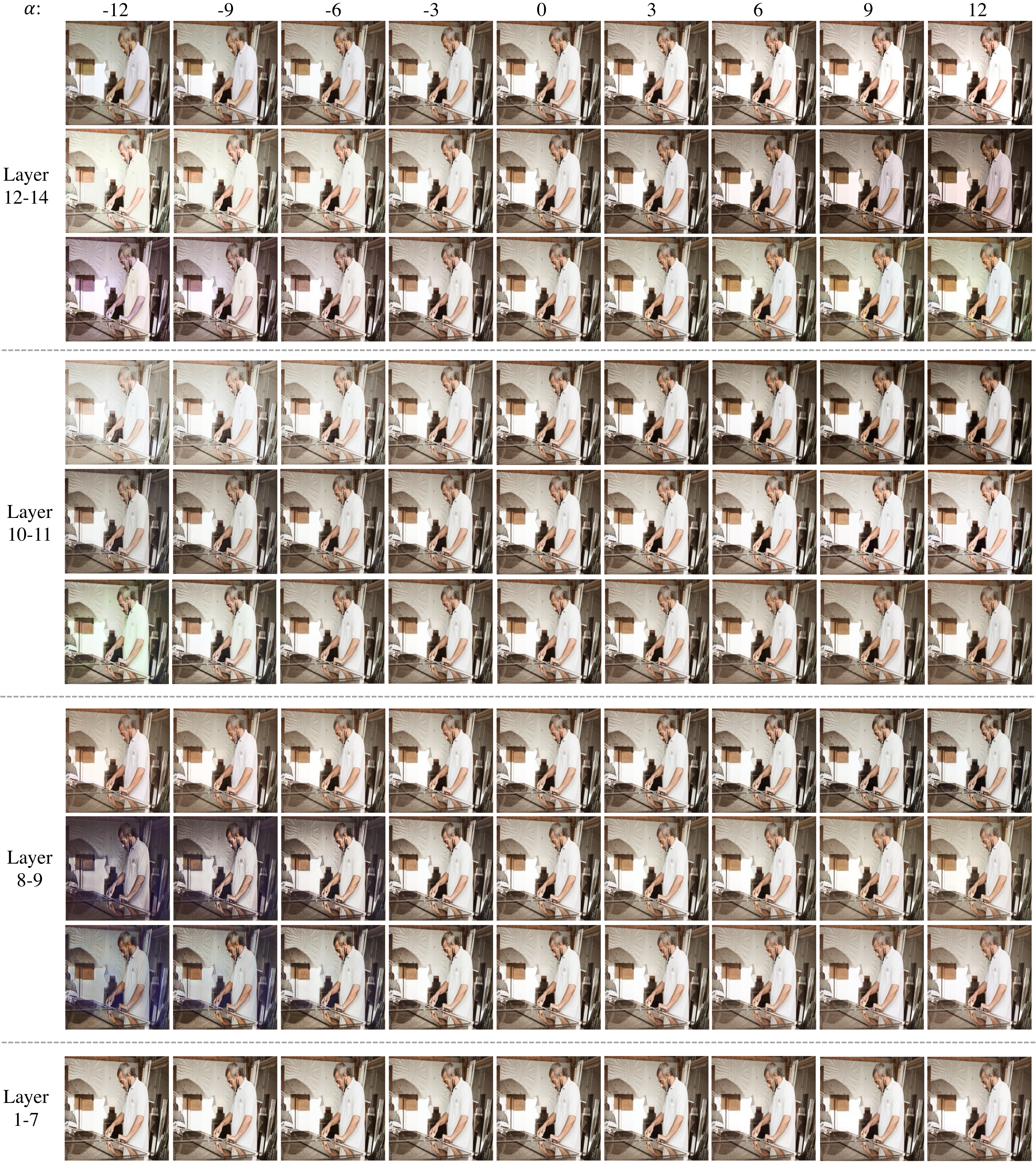}
    \caption{The visualization of the SeFa disentanglement on different layers. We select top-3 principle directions in layer 12-14, 10-11, 8-9 and top-1 direction for layer 1-7.}
    \label{fig:layerwise_sefa2}
\vspace{-2mm}
\end{figure*}

We analyze the effect of $\mathbf{w}$ at different layers using SeFa~\cite{shen2021closed}.
We compute the principle directions of the $\mathcal{W}$ space from the parameters of the affine matrix in the designated layers.
For a given principle direction $\mathbf{n}$ and the $\mathbf{w}_0$ of the input image, we traverse the $\mathcal{W}$ space using a scalar $\alpha$ as 
\begin{equation}
    \mathbf{w} = \mathbf{w}_0 + \alpha \mathbf{n}.
\end{equation}
Here we analyze the 256x256-resolutional generator with 14-layer decoder, where 12-14 conv-layers have output feature map size of 256x256, 10-11 conv-layers 128x128, 8-9 conv-layers 64x64, and 1-7 conv-layers have the size from 4x4 to 32x32.
Fig.~\ref{fig:layerwise_sefa1} and \ref{fig:layerwise_sefa2} show two examples for the traversing at these layers.
We can see that in the high-level layers, the traverse of $\mathbf{w}$ exhibits salient color change, while for the low-level layers (layer 1-7), it does not.
This means that the $\mathbf{w}$ in low-level layers will be ignored by the generator.
And at different high-level layers, they seems to be able to achieve the similar effect, such as the green effect can be achieved by all the 12-14, 10-11, 8-9 layers in Fig.~\ref{fig:layerwise_sefa1}, so it is still not quite clear to us what editing styles different layers emphasize.
And this could be further studied for future work.
Moreover, for different images, the same direction generally has the same semantic according to the comparison of Fig.~\ref{fig:layerwise_sefa1} and \ref{fig:layerwise_sefa2}.
However, the same style effect will overlay the original color style of the image, which explains why the final styles of the two examples have little different.

\section{More visual results}
\subsection{Retrieval}\label{appx:retrieval}
\begin{figure*}[t]
	\centering
	\includegraphics[width=1\linewidth]{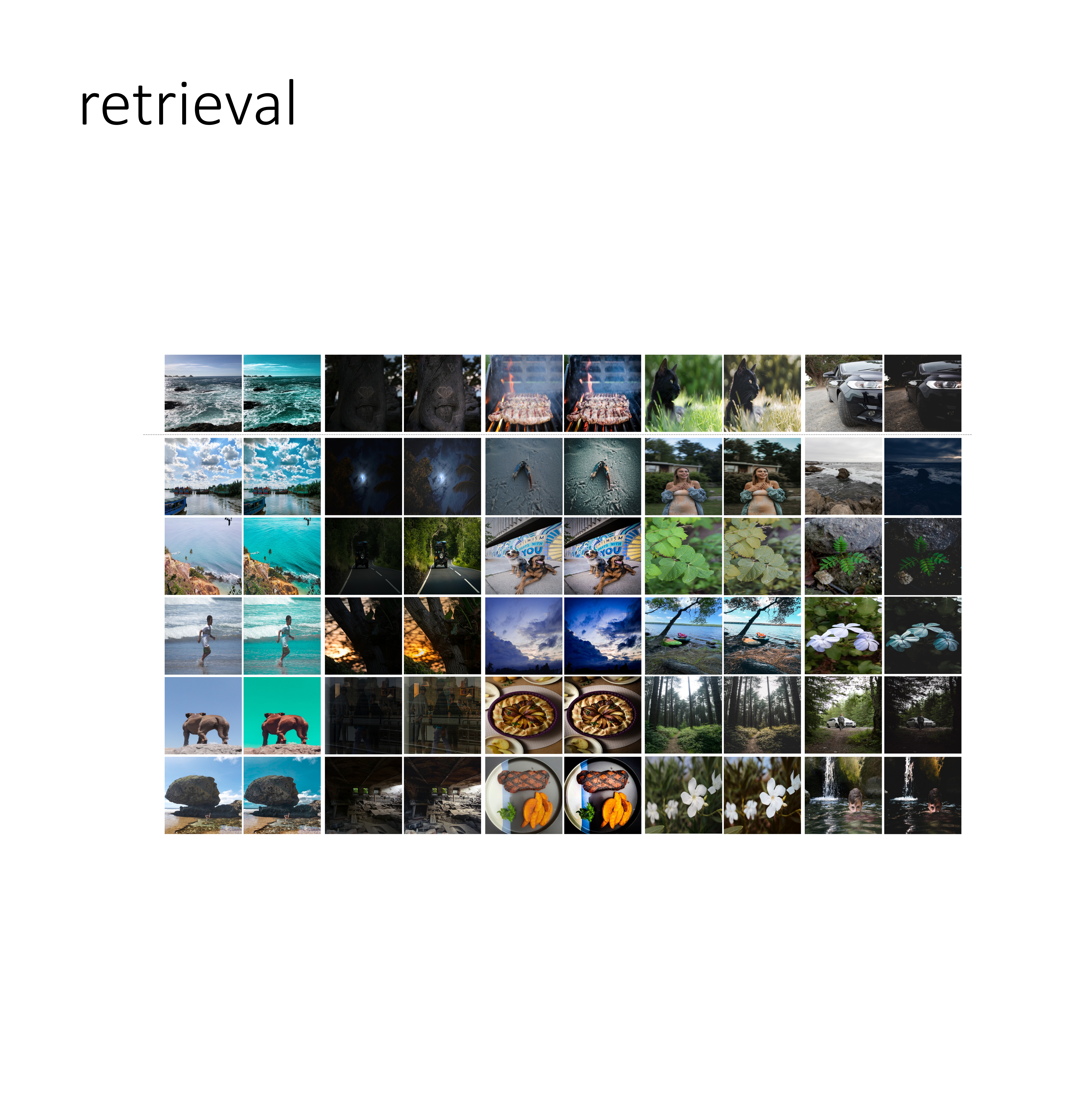}
    \caption{The visualization of the image pair retrieval results. The first row is the query pair, and the second to the last row are five retrieved pairs. For each pair, the left is source and the right is target.}
    \label{fig:appx_retrieval}
\vspace{-2mm}
\end{figure*}

Examples are shown in Fig~\ref{fig:appx_retrieval}.
\subsection{Cluster}
\begin{figure*}[t]
	\centering
	\includegraphics[width=1\linewidth]{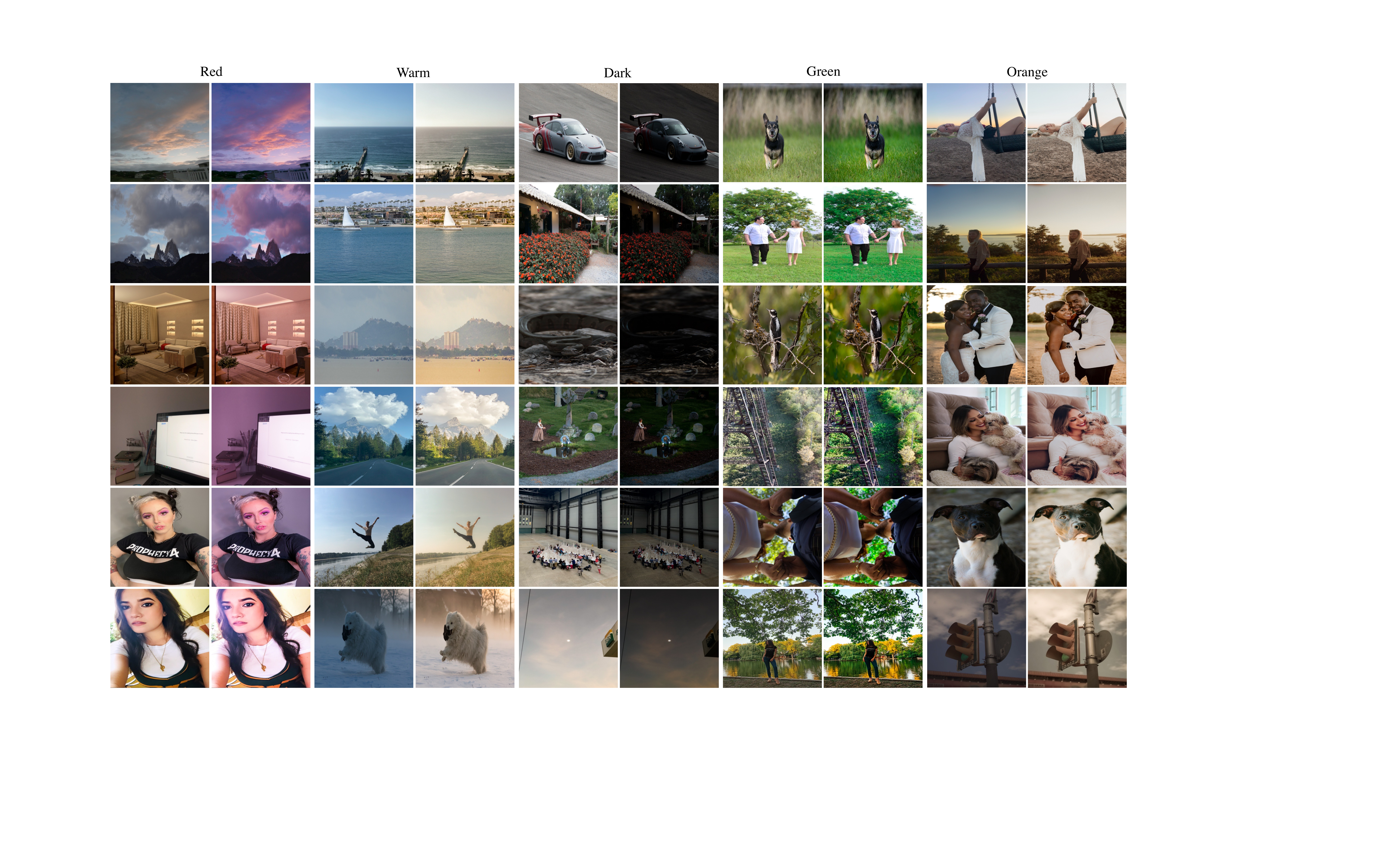}
    \caption{The visualization of the cluster results. For each image pair, the left is source and the right is target.}
    \label{fig:appx_cluster}
\vspace{-2mm}
\end{figure*}

More examples are shown in Fig.~\ref{fig:appx_cluster}.
\subsection{Multimodal Image Editing}
\begin{figure*}[t]
	\centering
	\includegraphics[width=1\linewidth]{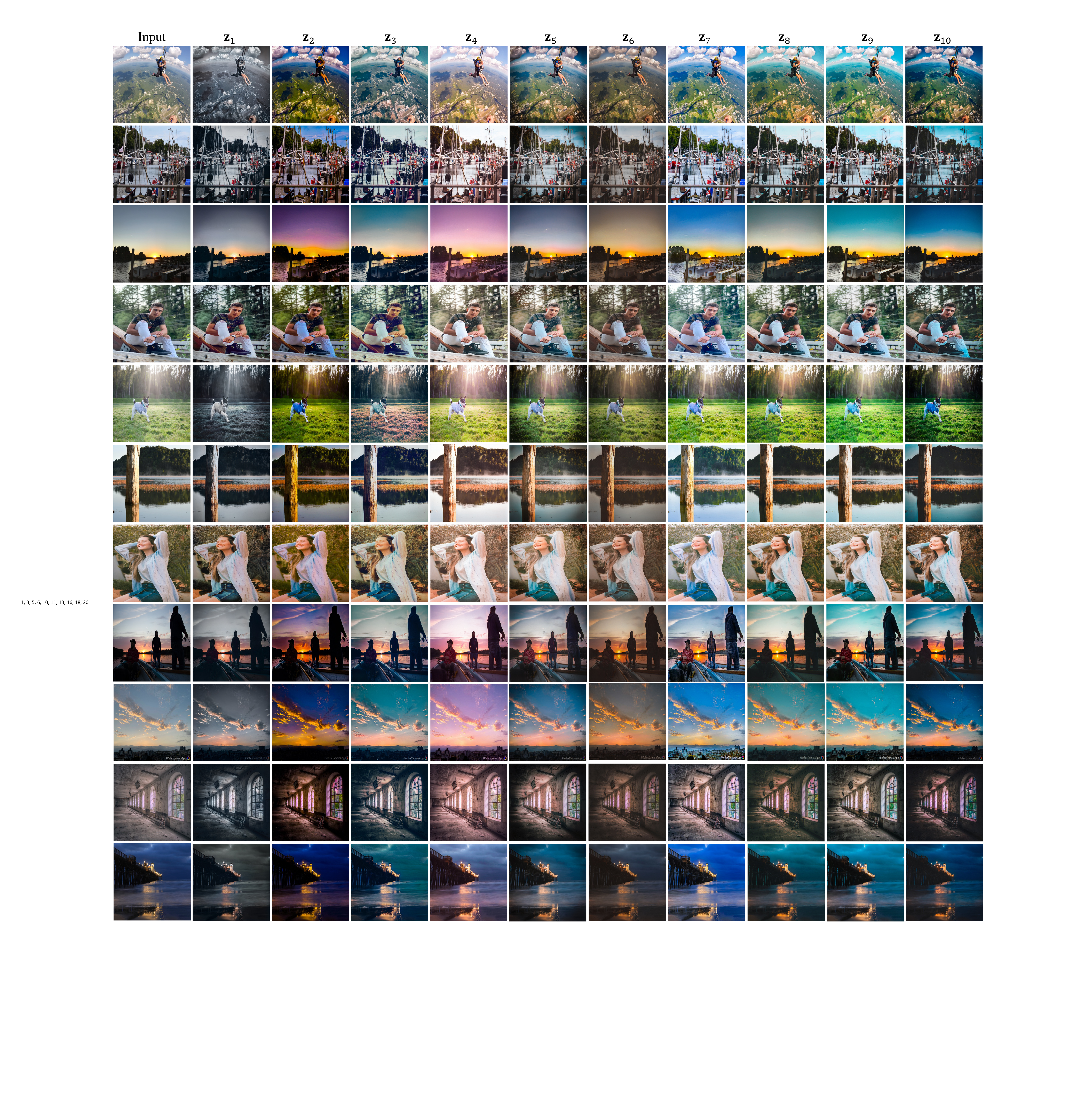}
    \caption{The visualization of the multimodal image editing result. Each column corresponds to the same $\mathbf{z}$, indicating one $\mathbf{w}$ has globally the same editing effect for all the images.}
    \label{fig:appx_multimodal}
\vspace{-2mm}
\end{figure*}

More examples are shown in Fig.~\ref{fig:appx_multimodal}.
\subsection{Exemplar-Based Image Editing}
\begin{figure*}[t]
	\centering
	\includegraphics[width=1\linewidth]{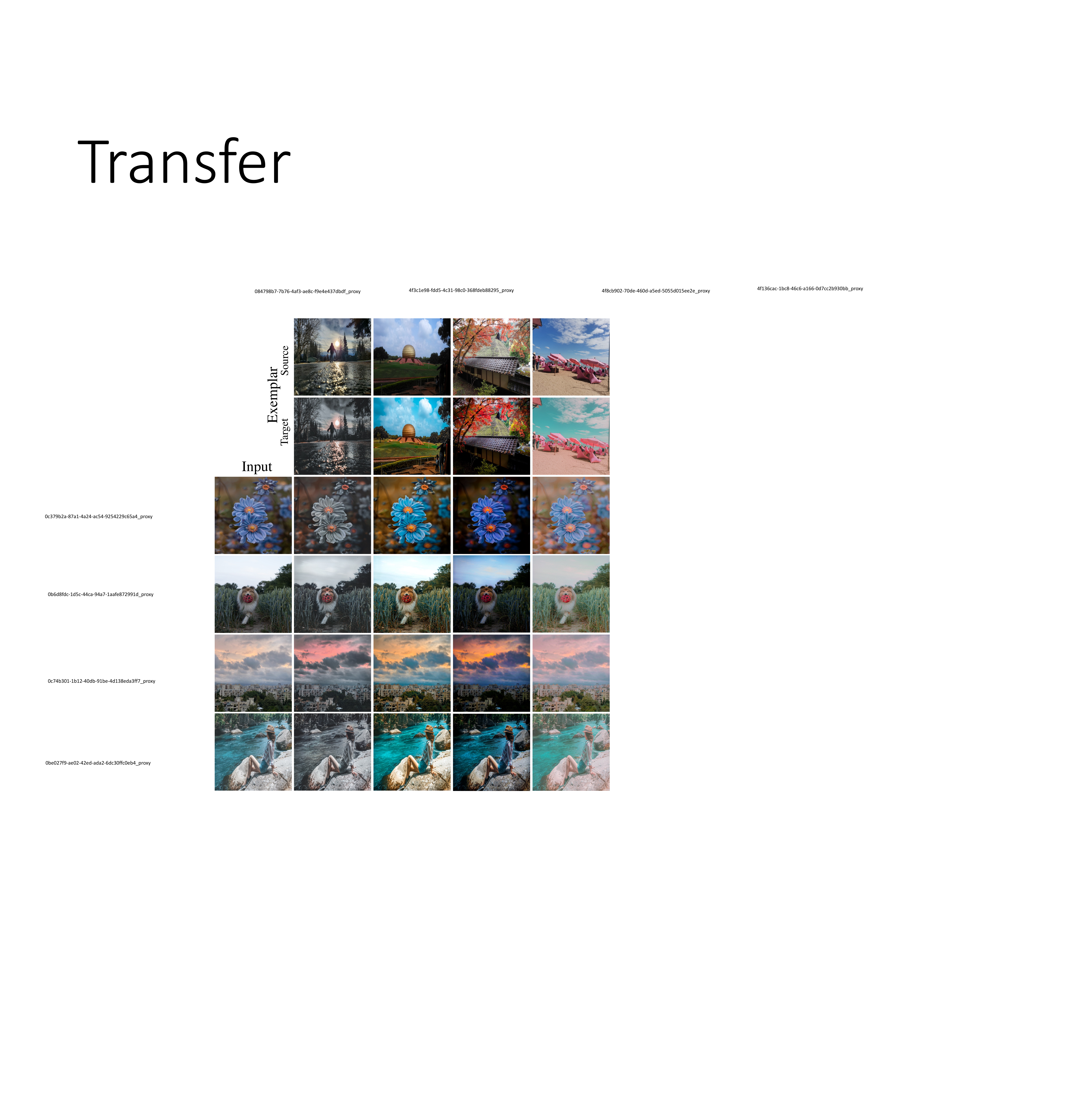}
    \caption{The visualization of exemplar-based image editing.}
    \label{fig:appx_transfer}
\vspace{-2mm}
\end{figure*}

More examples are shown in Fig.~\ref{fig:appx_transfer}.
\subsection{Language-Guided Image Editing}
\begin{figure*}[t]
	\centering
	\includegraphics[width=1\linewidth]{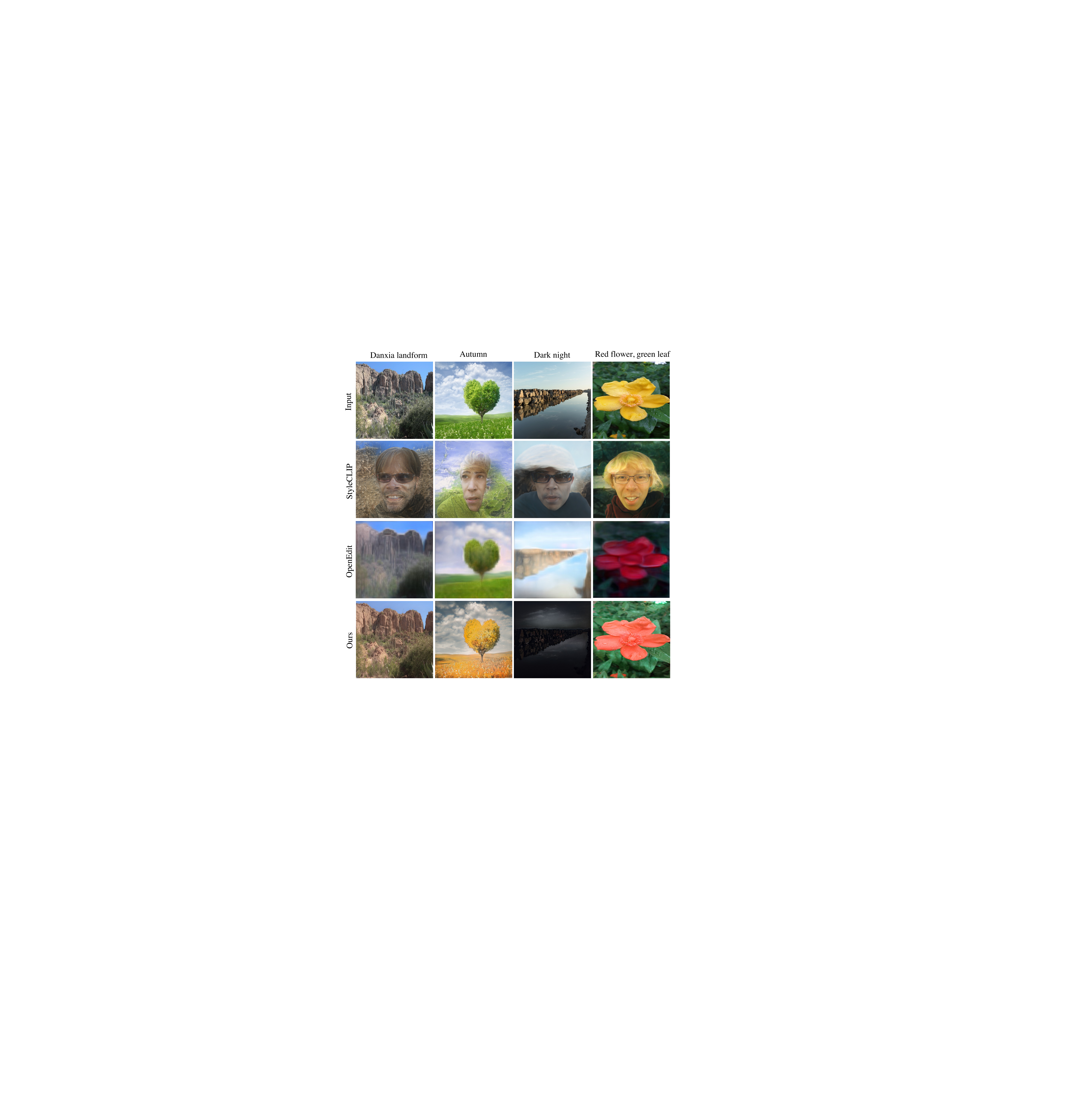}
    \caption{The visualization of the zero-shot LGIE.}
    \label{fig:appx_clip}
\vspace{-2mm}
\end{figure*}

More examples for zero-shot LGIE are shown in Fig.~\ref{fig:appx_clip}. Note that the flower example does not receive mask input, but it still can handel local editing. This verifies that our model can understand the language semantic and the generator has good ability for local editing.

\section{Language-Guided Image Editing}
\subsection{Supervised LGIE} \label{appx:supervised_LGIE}
We further provide more detailed introduction of the dataset, metrics, and more comprehensive comparison with other methods collected from~\cite{shi2021learning}.

\noindent\textbf{Dataset.} 
\textit{MA5k-Req.} MA5k-Req~\cite{shi2021learning} augments the language request to the image pairs in the MIT-Adobe FiveK dataset~\cite{bychkovsky2011learning}. 
It contains 24,750 image pairs with one language annotation each and is divided into 17,325/2,475 /4,950 for train/val/test split.

\noindent\textbf{Metrics.}
We follow the metrics in ~\cite{shi2021learning}.
\begin{itemize}[noitemsep, topsep=0pt]
\item \textit{L1} distance directly measures the averaged pixel absolute difference between the generated image and ground truth image with pixel normalized to 0-1.
\item \textit{SSIM} measures image similarity through luminance, contrast, and structure.
\item \textit{FID} measures the Fréchet distance between two Gaussians fitted to feature representations of the Inception network over the generated image set and ground truth image set. 
\item \textit{Image variance} $\sigma$ measures the language controllability by computing the pixel variance of 10 output of the same input image controlled by different languages.
\end{itemize}

\noindent\textbf{Comparison methods.}
\begin{itemize}[noitemsep, topsep=0pt]
\item \textit{Input}: the evaluation between input and target image.
\item \textit{Bilinear GAN}~\cite{mao2019bilinear}, \textit{SISGAN}~\cite{dong2017semantic}, \textit{TAGAN}~\cite{nam2018text}: these three methods are trained  by learning the mapping between the caption and image without image pairs. Since there is not image caption in our task but the paired image and request, we drop the procedure of image-caption matching learning but adapt them with the L1 loss between input and target images.
\item \textit{Pix2pixAug}~\cite{wang2018learning}: the pix2pix model~\cite{isola2017image} augmented with language used in \cite{wang2018learning}.
\item \textit{GeNeVa}~\cite{el2019tell}: a GAN-based dialogue guided image editing method. We use it for single-step generation.
\item \textit{RL}: an RL approach introduced in \cite{shi2021learning}.
\item\emph{T2ONet}~\cite{shi2021learning}: T2ONet map the language request to a series of editing operations using weak supervision.
\item\emph{EDNet}~\cite{jiang2021language}: EDNet enforce the language controllability using cyclic loss.
\end{itemize}

\begin{table}[t]\centering
\ra{1.2}
\scalebox{0.95}{
\begin{tabular}{@{}lrrrr@{}}
\toprule
\cmidrule{2-5} 
& L1 $\downarrow$ & SSIM$\uparrow$ & FID$\downarrow$ & $\sigma_{\times 10^2}$ $\uparrow$\\
\midrule
Target &  -   &   -    &   -     & -    \\
Input & 0.1190 & 0.7992 & 12.3714 & - \\
Bilinear GAN~\cite{mao2019bilinear} & 0.1559 & 0.4988 & 102.1330 & 0.8031 \\
Pix2pixAug~\cite{wang2018learning} & 0.0928 &  0.7938 & 14.5538 & 0.5401 \\
SISGAN~\cite{dong2017semantic} & 0.0979 & 0.7938 & 30.9877 & 0.1659 \\
TAGAN~\cite{nam2018text} & 0.1335 & 0.5429 & 43.9463 & 1.5552 \\ 
GeNeVa~\cite{el2019tell} & 0.0933 & 0.7772 & 33.7366 & 0.6091 \\
RL~\cite{shi2021learning} & 0.1007 & 0.8283 & 7.4896 & \textbf{1.6175}  \\
T2ONet~\cite{shi2021learning} & 0.0784 & 0.8459 & 6.7571 & 0.7190 \\
EDNet~\cite{jiang2021language} & - & - & 9.9500 & -  \\
Ours & \textbf{0.0731} & \textbf{0.8721} & \textbf{5.9791} & 0.6809  \\
\bottomrule
\end{tabular}}
\caption{Quantitative results on MA5k-Req test sets. $\sigma_{\times 10^2}$ means that the image variance has been scaled up 100 times.}
\label{tab:LGIEResult}
\end{table}

\section{Data Collection}\label{appx:data_collection}
We collect the dataset called Discover-Req, where we augment the language request that describes what are edited for the before- and after-images.
The whole process obtains the permit and the Discover images are allowed for research use.
Totally we collected the language annotation for 4423 pairs of images with one sentence from Photoshop expert and three sentences from amateurs for each pair.
The expert are hired from Upwork\footnote{https://www.upwork.com/} and the amateurs from ScaleAI\footnote{https://scale.com/}.
The annotation quality of the expert is trustable. 
To control the quality of amateurs, we only hire those who pass the annotation test, and the annotation result must be approved by another worker. 

\section{Tag List creating}\label{appx:tag_list}
The final tag list is: \emph{dark, blue, red, white, vivid, vintage, warm, brown, clear, clarity, green, natural, yellow, orange, retro, cool, black, vignette, vibrant.}

The steps for creating this tag list is as follows.
We firstly create a prior tag list based on the Adobe Photoshop commonly used style effect.
Next, we tokenize all the annotated sentences in Discover-Req dataset, stemitize all the tokens, and manually select the style-like tokens and merge them with the prior tag list.
Then we remove the tag that occurs to most of the image such as \emph{bright} ``contrast''.
Finally, we filter out those tags that occur less than 5 times among all sentences. 

\section{Customized Purity}\label{appx:purity}
Standard purity is computed for single labeled sample.
However, each of our image pair has been labeled with multiple tags (the tokenized sentence may contain multiple valid tags).
Therefore, we will extend the computation for purity to support multi-label situation.
Specifically, for each cluster $C_i$, we firstly construct its corresponding tag pool $T_i$ by collecting all the tag labels of all the samples in this cluster (the tag pool allows the same tag to occur many times).
Next, for each tag $t_j$ in the tag list of length $L$, we count $t_j$ in each $T_i$ and find the cluster with the maximum count of $t_j$ as $C_j$.
So now we have assigned the tag $t_j$ to $C_j$.
Note that in this way, one cluster might be assigned by multiple tags, but it does not matter.
Then, we count the number of $t_j$ in $C_j$ as $N_j$ and let $|\cdot|$ denote the total number of the elements of a set, the purity is defined as
\begin{equation}
    \mathrm{purity} = \frac{\sum_j^L N_j}{\sum_j^L |C_j|}.
\end{equation}


\end{document}